%% file: neurips_2019.tex
\documentclass{article}

     \usepackage[preprint]{neurips_2019}

\usepackage[utf8]{inputenc} 

\usepackage{biblatex}
\usepackage[T1]{fontenc}    
\usepackage{hyperref}       
\usepackage{url}            
\usepackage{booktabs}       
\usepackage{amsfonts}       
\usepackage{nicefrac}       
\usepackage{microtype}      

\usepackage{times}
\usepackage{latexsym}
\usepackage{helvet}  
\usepackage{courier}  
\usepackage{graphicx, graphics}
\usepackage{amssymb}
\usepackage{pgfplots}
\usepackage{mwe}

\usepackage{color}
\usepackage{multirow}
\usepackage{bbm}
\usepackage{algorithm,algorithmic}
\usepackage{float}
\usepackage{caption, subcaption}
\usepackage{wrapfig}
\usepackage{amsmath}
\usepackage{lscape}
\usepackage{lipsum}
\usepackage{tabularx}

\newcolumntype{P}[1]{>{\centering\arraybackslash}p{#1}}

\newcommand{\cev}[1]{\reflectbox{\ensuremath{\vec{\reflectbox{\ensuremath{#1}}}}}}

\title{Visual Relationship Detection using Scene Graphs: \\ A Survey}

\author{%
  Aniket Agarwal\thanks{The first three authors have contributed equally and the names are arranged in an alphabetical order.}  \\
  {\tt\small aagarwal@ma.iitr.ac.in}
  \And
  Ayush Mangal\footnotemark[1]  \\
  {\tt\small amangal@cs.iitr.ac.in}
  \And
  Vipul\footnotemark[1] \\
  {\tt\small vipul@ee.iitr.ac.in}
  \and \\
  Vision and Language Group\\ 
  Indian Institute of Technology Roorkee \\
  {\tt\small vlgiitr.github.io}
}

\begin{document}

\maketitle

\begin{abstract}
  Understanding a scene by decoding the visual relationships depicted in an image has been a long studied problem. While the recent advances in deep learning and the usage of deep neural networks have achieved near human accuracy on many tasks, there still exists a pretty big gap between human and machine level performance when it comes to various visual relationship detection tasks. Developing on earlier tasks like object recognition, segmentation and captioning which focused on a relatively coarser image understanding, newer tasks have been introduced recently to deal with a finer level of image understanding. A Scene Graph is one such technique to better represent a scene and the various relationships present in it. With its wide number of applications in various tasks like Visual Question Answering, Semantic Image Retrieval, Image Generation, among many others, it has proved to be a useful tool for deeper and better visual relationship understanding. In this paper, we present a detailed survey on the various techniques for scene graph generation, their efficacy to represent visual relationships and how it has been used to solve various downstream tasks. We also attempt to analyze the various future directions in which the field might advance in the future. Being one of the first papers to give a detailed survey on this topic, we also hope to give a succinct introduction to scene graphs, and guide practitioners while developing approaches for their applications.
\end{abstract}


\input{introduction}
\input{building_blocks}

\input{problem_formulation}
\input{methods}

\input{datasets}

\input{applications}
\input{comparison}

\input{future_directions}

\input{conclusion}

\subsubsection*{Acknowledgments}
We thank the invaluable inputs and suggestions by Dakshit Agrawal, Aarush Gupta and other members of Vision and Language Group, IIT Roorkee, that were integral for the successful completion of this project. We would also like to thank the Institute Computer Center (ICC) IIT Roorkee for providing us with computational resources.

\medskip
\printbibliography


\end{document}

%% file: introduction.tex
\section{Introduction}
\label{introduction}



\subsection{Motivation}
The past decade has seen a steady increase in the usage of deep learning techniques for solving various tasks, which can be credited to the presence of huge datasets like Imagenet \cite{deng2009imagenet}, MS-COCO \cite{lin2014microsoft}, etc. for training these data-driven models. With the breakthrough results of AlexNet \cite{krizhevsky2012imagenet} in the ILSVRC challenge \cite{russakovsky2015imagenet}, deep neural networks have become a new standard over the traditional methods that used handcrafted shallow features, to get better accuracy on various tasks owing to their powerful ability to learn different levels of features and powerful representations from the input data. Since then, deep neural networks have been used to achieve SOTA (State of the Art) on various tasks such as Image Recognition \cite{simonyan2014very, szegedy2015going, he2016deep, huang2017densely}, Object Recognition \cite{girshick2014rich, girshick2015fast, ren2015faster, redmon2016you} and Image Segmentation \cite{ronneberger2015u, zhao2017pspnet, chen2017rethinking, chen2018encoder}, to name a few. It is clear that the ultimate aim of all these tasks has been to learn even more powerful representations from the input data by solving even more challenging tasks like Image Captioning \cite{xu2015show, karpathy2015deep, lu2018neural}, Visual Question Answering \cite{antol2015vqa, lu2016hierarchical, anderson2018bottom} and  Image generation \cite{NIPS2014_5423, kingma2013auto, dinh2016density}.

\begin{figure}[t]
\captionsetup{labelfont=bf}
    \includegraphics[width=8cm]{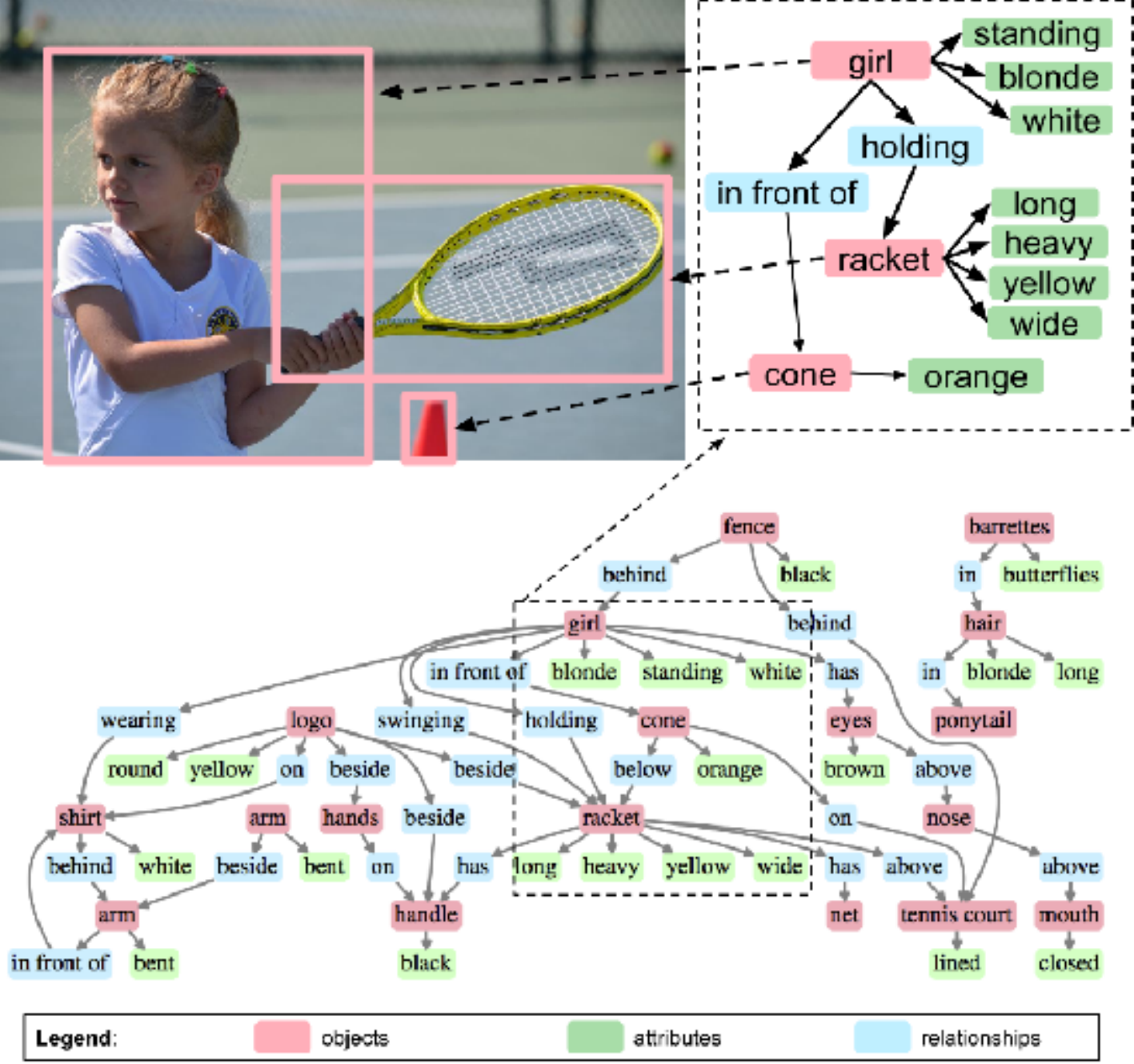}
\centering
\caption{An example of a scene graph taken from \cite{johnson2015image}. The scene graph encodes objects ("girl"), attributes ("girl is blonde") and relationships ("girl holding racket").}
\label{fig:scene_graph1}
\end{figure}

As a result, to describe the image features and object relationships in an even more explicit and structured way, Scene Graphs \cite{johnson2015image} were proposed. Scene graphs capture the detailed semantics of visual scenes by explicitly modeling objects along with their attributes and relationships with other objects. With the nodes representing the objects in the scene and the edges linking them representing the relationships between the various objects in a dynamic graphical structure, we get a rich representation of the given scene in an image. An example for the same can be seen in Fig. \ref{fig:scene_graph1}, where the objects, their attributes and the relationships are being represented in a graphical way.

However, such rich representational power does come with an evident challenge of its generation, a problem we are going to delve into much deeper in further sections. The pipeline for Scene Graph Generation (SGG) \cite{xu2017scene, li2017scene, li2018factorizable} can be broken down into several steps: object detection for node creation, graph generation using the created nodes and an iterative updation of the relationship and node features to obtain the desired scene graph. The interdependence of the subcomponents of the pipeline for the task of scene graph generation makes it a much more complicated task than the individual tasks of object detection, graph creation, etc. Hence for a comprehensive understanding of the task, the entire pipeline would be taken up in detail along with the various problems associated with each of its steps in further sections. We will also be discussing some of the major reasons for the limited usage of scene graphs for various downstream tasks, despite their strong scene representation power.

As a result, we will be having most of our discussion on the various techniques developed in the past literature to tackle the above stated problems. Apart from these, we will also examine the various applications of scene graphs and how they can be used to solve various downstream tasks such as Image Captioning, VQA, Scene Generation, Semantic Image Retrieval, etc. The rest of the paper can be divided into the following sections:

\begin{itemize}
    \item Section \ref{building_blocks} covers some common model architectures that would be frequently mentioned throughout the paper.
    \item Section \ref{problem_formulation} defines Scene Graph and briefs about the general pipeline used for its generation.
    \item Section \ref{methods} jumps into the taxonomy used for classifying the various models in the past literature and also comprehensively describes the methodology used for the SGG task through the years.
    \item Section \ref{datasets} briefs about some common datasets used in scene graph generation and related tasks.
    \item Section \ref{applications} discusses the  applications of scene graphs for solving various downstream tasks.
    \item Section \ref{performance} gives a comparative study on the effectiveness of the various techniques discussed.
    \item Section \ref{future_directions} explores the possible future trends in research and applications of this field.
\end{itemize}

To the best of our knowledge, this is the \emph{first survey paper} giving a comprehensive overview of the methodologies and terminologies used in Scene Graph research. While there has been some survey work done on the various visual relationship detection tasks \cite{wu2017visual, chaudhari2019attentive}, we are the first ones to target the Scene Graph Generation problem and how it has been useful for visual relationship tasks. We also give the \emph{first taxonomy} to classify the work done in this field, which we believe would help to streamline the future work in this area. Lastly, we also discuss upon the past trend and \emph{future directions} for research in this field. With such wide applicability, we feel this comprehensive evaluation of the past work is required for further advancement in this field, and at the same time give the required exposure to someone starting in this field.

\subsection{Term Definition}
\label{term_definition}

To make this survey easier to read, we would first like to define some basic terms/problem tasks that are constantly used in the literature of scene graphs.

\begin{enumerate}
    \item \textbf{Object Detection:}  The problem involves simultaneously classifying and localizing objects using a bounding box. Although recent works \cite{redmon2018yolov3} show superior performance, we will be restricting our study to Faster RCNN \cite{ren2015faster} (explained in the next section) because of its frequent usage in the scene graphs literature.
    
    \item \textbf{Image Captioning:}  The problem revolves around generating a caption \cite{anderson2018bottom, gu2019unpaired} that can summarize the scene represented in the image along with the relationships among the objects.
    
    \item \textbf{Visual Question Answering:}  As described in \cite{antol2015vqa}, given an image and a natural language question about the image, the task is to provide a natural language answer. Mirroring real-world scenarios, both questions and answers are open-ended.
    
    \item \textbf{Semantic Image Retrieval:}  The image retrieval task \cite{johnson2015image, schuster2015generating} is to retrieve an image from a database by describing their contents, while the semantic part arises from the fact that not only are we specifying the "objects" in the image for the retrieval, but also structured relationships and attributes involving these objects.
    
    \item \textbf{Visual Grounding of Scene Graph:}  Frequently used in the semantic image retrieval task, it involves associating a given scene graph to an image. Formally if an image is represented by a set of bounding boxes $B$, a grounding of a scene graph $G = (O, E)$, where $O$ represents objects and $E$ represents relationships in the given scene graph, is a map $\gamma : O \rightarrow B$.
    
    \item \textbf{Image Generation:}  The image generation task is a generative task where our model, when once trained on the training data, is able to approximate the training distribution and hence sample from random variables of the distribution to "generate" more data pertaining to the same distribution.
\end{enumerate}

%% file: building_blocks.tex
\section{Building Blocks}
\label{building_blocks}

To make the readers acquainted with the common models/architectures that are used in the scene graph literature, we will be defining various common models here in detail to give a better insight into the processes that are carried out in the whole pipeline.

\subsection{Faster RCNN}
Faster RCNN \cite{ren2015faster} is a successor of the RCNN \cite{girshick2014rich} family. It has both improved speed and accuracy over RCNN and Fast RCNN \cite{girshick2015fast}. The main contribution of Faster RCNN is the introduction of \emph{Region Proposal Network (RPN)}, which has become the de-facto choice for generating region proposals over the years. From R-CNN to Faster RCNN, most individual blocks of an object detection system like region proposals, feature extraction, bounding box regression etc. have been gradually integrated into a unified, end-to-end learning framework.  Selective Search, which was used earlier for object proposals, is serviceable but takes a lot of time and sets a bottleneck for speed as well as accuracy. Therefore a better detector could surely increase the speed and efficiency.  Several works \cite{zhou2015learning,zhou2014object,7420739} have shown that deep ConvNets have a remarkable ability to localize objects in the convolutional layers, an ability which is weakened in the fully connected layers. Faster RCNN utilizes this property and hence uses weight sharing among the convolutional layers to jointly propose regions and detect objects correctly in an end-to-end manner rather than having object proposals generated as a pre-processing step. RPNs in the Faster RCNN model uses the same feature encoder as Fast RCNN and shares the last convolution layer. To predict regions of multiple scales and aspect ratios, RPNs use the novel concept of anchors. Although it breaks the bottleneck, there is still a lot of computation redundancy, which was further dealt with in later works such as RFCN \cite{dai2016rfcn} and light head RCNN \cite{li2017lighthead}.
\subsection{Graph Neural Networks (GNN)}
A graph is a non-euclidean datatype, and to reason over it, Graph Neural Network (GNN) \cite{4700287} is a common approach nowadays. The primary challenge in this domain is finding a way to represent, or encode, graph structure so that machine learning models can easily exploit it. The traditional approaches were based upon user-defined heuristics to encode the structural information about a graph. In contrast, the recent work automatically learns to encode the same using deep neural networks and dimensionality reduction techniques. This encoded information can be further utilized for task-specific objectives such as node classification, link prediction, or clustering. Scene Graph, being a graph-based structure, uses several GNN techniques for encoding structural information for better feature generation. GNNs focus on exploiting the available information in the form of a graph, whereas the Scene Graph Generation task tries to produce one.


\begin{figure}[]
\captionsetup{labelfont=bf}
\centering
\begin{subfigure}[b]{0.32\textwidth}
\includegraphics[width=1.5in]{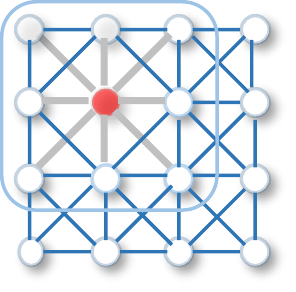}
\caption{2D Convolution. Analogous to a graph, each pixel in an image is taken as a node where neighbors are determined by the filter size. The 2D convolution takes the weighted average of pixel values of the red node along with its neighbors. The neighbors of a node are ordered and have a fixed size.}
\end{subfigure}%
\hspace{2cm}
\begin{subfigure}[b]{0.32\textwidth}
\includegraphics[width=1.5in]{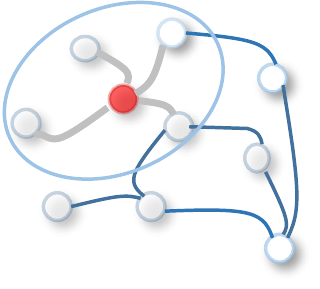}
\caption{Graph Convolution. To get a hidden representation of the red node, one simple solution  of the graph convolutional operation is to take the average value of the node features of the red node along with its neighbors. Different from image data, the neighbors of a node are unordered and variable in size.}
\end{subfigure}
\caption{2D Convolution vs. Graph Convolution \cite{wu2019comprehensive}.}
\label{gcn_cnn}
\end{figure}

\subsection{Graph Convolutional Networks (GCN)}
Convolutional graph neural networks (ConvGNNs) generalise the convolution operation from regular patterned grids like images to arbitrarily shaped graphs. They can be divided into two main categories: 1) \emph{Spatial GCNs} \cite{atwood2016diffusion,micheli2009neural,hamilton2017inductive}, which utilizes a node's spatial relations to define the convolution operation, updating the node's representation based on its neighbors; 2) \emph{Spectral GCNs} \cite{bruna2013spectral,defferrard2016convolutional,kipf2016semi}, which are based on a graph signal processing approach and considers convolution operation as the multiplication of a signal with a filter parameterized in the Fourier domain.

Improving upon the earlier works on spectral graph convolutions \cite{bruna2013spectral} and fast localized spectral filtering \cite{defferrard2016convolutional}, Kipf \& Welling \cite{kipf2016semi} introduced \emph{Graph Convolutional Network (GCN)} which is based on a localized first order approximation of \cite{defferrard2016convolutional}.

Graph Convolutional Network (GCN) \cite{kipf2016semi} proposes the following layer wise propagation rule:
\begin{equation}
  \textstyle
  H^{(l+1)}= \sigma\!\left(\tilde{D}^{-\frac{1}{2}} \tilde{A}\tilde{D}^{-\frac{1}{2}}H^{(l)} W^{(l)} \right) \, .
\label{eq:gcn-layer}
\end{equation}
Here $A$ is the adjacency matrix of the graph, $I_N$ is the identity matrix which is used to provide self connections in the new adjacency matrix $\tilde{A} = A + I_N$.  $\tilde{D}_{ii} = \sum_j \tilde{A}_{ij}$  is a diagonal matrix specifying node degrees of $\tilde{A}$ and is used in the re-normalization trick to avoid numerical instability and exploding/vanishing instability problem usually faced by deep networks. $W^{(l)}$ is a layer-specific trainable weight matrix. $\sigma(\cdot)$ represents an activation function. $H^{(l)}\in \mathbb{R}^{N\times D}$ is the matrix of activations for the $l^{\text{th}}$ layer, where $H^{(0)}=X$ (the input graph features).

 The various approaches for scene graph generation incorporate different variants of the above specified GCN framework to deal with the graphical nature of the problem, with a common approach being the usage of attention \cite{zhang2018gaan,yang2018graph} to find the relative importance of neighbours of a node for a subsequent updation of its features.
 
 Graph Attention Networks \cite{velivckovic2017graph} uses multi-headed self attention with the following propagation rule:
 
 \begin{equation}
	\vec{h}'_i =  \sigma\left(\frac{1}{K}\sum_{k=1}^K \sum_{j\in\mathcal{N}_i}\alpha_{ij}^k{\bf W}^k\vec{h}_j\right)
\end{equation}
 Where $\vec{h}'_i$ is the output representation of the node and $\vec{h}_j$ denotes the input representation of the node's neighbours. ${\bf W}^k$ is an input linear transformation weight matrix and $\alpha_{ij}^k$ is the normalised attention score calculated by the k-th attention head using the formula :
\begin{equation}
\vspace{0.1cm}
\alpha_{ij} = \mathrm{softmax}_j(e_{ij})=\frac{\exp(e_{ij})}{\sum_{k\in\mathcal{N}_i} \exp(e_{ik})}.
\end{equation}
where $e_{ij}$
\begin{equation} 
e_{ij} = a({\bf W}\vec{h}_i, {\bf W}\vec{h}_j)
\vspace{0.15cm}
\end{equation}
gives the relative importance between the nodes $i$ and $j$ using an attention mechanism $a$ which is a single layer feed forward network parameterised by a weight vector $\vec{\bf a}$.

%% file: problem_formulation.tex
\section{Scene Graph Defintion}
\label{problem_formulation}

Before proceeding further, we would like to formally define the term \emph{Scene Graph} and the various components associated with it.

A \emph{Scene Graph} is a graphical data structure that describes the contents of a scene. A scene graph encodes object instances, attributes of objects, and relationships between objects. Given a set of object classes $\mathcal{C}$, a set of attribute types $\mathcal{A}$, and a set of relationship types $\mathcal{R}$, we define a scene graph \emph{G} to be a tuple \emph{G = (O, E)} where $O = (o_{1},..., o_{n})$ is a set of objects and $E \subseteq O \times \mathcal{R} \times O$ is a set of edges. Each object has the form $o_i = (c_i, A_i)$ where $c_i \in \mathcal{C}$ is the class of the object and $A_i \subseteq \mathcal{A}$ are the attributes of the object. \textbf{In the subsequent sections, we will be following the above-mentioned notations/symbols for the various components of a scene graph unless stated otherwise.}

We will now outline the general pipeline followed by most of the previous works to give the readers a basic understanding of the various components of the Scene Graph Generation process. A diagrammatic representation for the whole process can also be seen in Fig. \ref{fig:pipeline}.

\subsection{Region Proposal Module}
The primary purpose of this module is to propose multiple objects that are identifiable within a particular image. RPN \cite{ren2015faster} is one of the default choices for this module in almost all the scene graph generation papers. The RPN quickly and efficiently scans every location in the image in order to assess whether further processing needs to be carried out in a given region. For this purpose, it outputs k bounding box proposals per image, each with an associated object confidence score. Afterwards, RPN ranks region boxes (called anchors) and proposes the ones most likely containing objects. Once we have the region proposals, they can be used further for graph construction and feature extraction. The module can be seen as represented in Fig. \ref{fig:pipeline}(a).

\subsection{Graph Construction Module}
Graph Construction Module is hidden in most of the scene graph generation networks, with its main work being to initialize the relationship features between the object pairs, with one of the common choices for this initialization being the usage of a Fully Connected Graph. Here Fully Connected Graph means every object proposal is considered as related to another object proposal, even though some relations like \emph{tire and a car} maybe more prominent than others like \emph{tire and a building}. Once the initial graph is created with some dummy relations between the detected objects, we move towards refinement of the node and edge features using the Feature Refinement Module. The Graph Construction Module can be seen as represented in Fig. \ref{fig:pipeline}(b).

\subsection{Feature Refining Module}
\begin{figure}[t]
\captionsetup{labelfont=bf}
    \includegraphics[width=14cm]{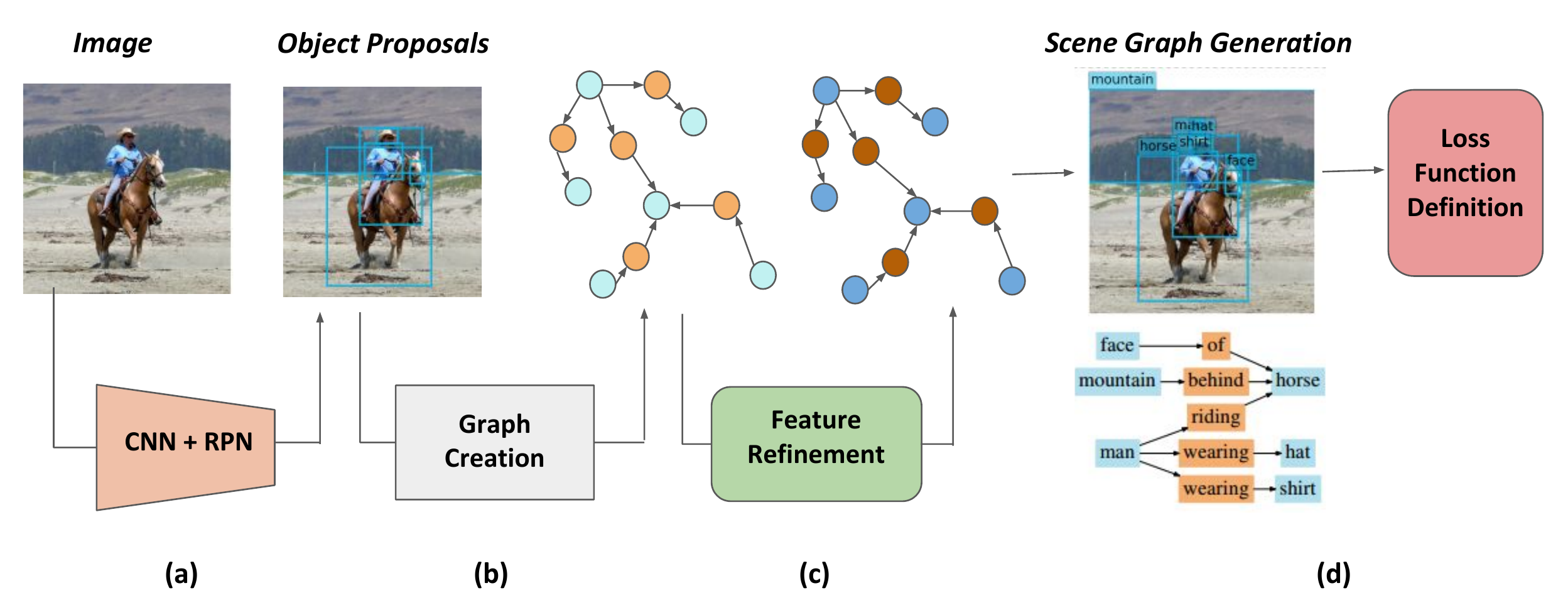}
\centering
\caption{A basic pipeline for the scene graph generation process, adapted from \cite{xu2017scene}. Given an image, an RPN block (a) is used to extract the object proposals along with their features, that are used to create a candidate graph (b). After the generation of a candidate graph, a feature refining module (c) is used to refine the features further. Once the refinement is done, a scene graph (d) is inferred according to the node and edge features. Subsequently, the model is trained using a loss function on the dataset.}
\label{fig:pipeline}
\end{figure}

Feature Refining Module is the most important and extensively studied module in Scene Graph Generation. Various methods \cite{xu2017scene,li2017scene,gay2018visual,chen2019knowledge,li2018factorizable,gay2018visual} were introduced from time to time to leverage its full capacity. The idea is to incorporate contextual information either explicitly or implicitly so that the detection process for objects and relations becomes more context-dependent. The intuition behind feature refining is the superior dependencies among \emph{<object-predicate-object>} triplet, i.e. if one object is "boy" and the other is "shirt," there is a high chance of "wear" as a predicate. Once the features are refined, the final graph is inferred and subsequently loss functions are defined to train the model. The Feature Refinement Module can be seen as represented in Fig. \ref{fig:pipeline}(c).

\subsection{Scene Graph Inference and Loss Functions}
With the node and edge features refined, we move onto inferring the final scene graph, which encodes the visual relationships in our scene. The inference task basically transforms the refined features to one-hot vectors for the object and predicate choices in the training dataset. Once the scene graph is obtained, a loss function is defined which usually calculates the likelihood between the obtained object-relation labels and the true labels in the training dataset. This module is represented in Fig. \ref{fig:pipeline}(d).

%% file: methods.tex
\section{Taxonomy of Scene Graph Generation Methods}
\label{methods}

Scene Graph Generation, being a relatively new research field, hasn't seen many enormous breakthroughs till now. Hence rather than being discrete, various methods were built on each other. Therefore dividing the various SGG techniques into discrete categories, like in the case of GNNs \cite{wu2019comprehensive}, is not suitable. Rather than concrete structural changes in the model architecture, most of the research in this field can be classified in the terms of the problem that they are mainly addressing in the SGG task . The four problems in the task that the past literature has majorly targeted, thus becoming the basis of our classification, are \emph{Efficient Graph Features Refinement}, \emph{Efficient Graph Generation}, \emph{Long-tailed Dataset Distribution} and \emph{Efficient Loss Function Definition}. As summarized in Table \ref{tab:taxonomy}, many works have targeted two or more of the problems specified at the same time.

\subsection{Efficient Graph Features Refinement}
\label{class_feature_refinement}
\textbf{\emph{Motivation}} ~~ The graph feature refinement module constitutes one of the most important aspects in the SGG pipeline, and hence much of the research in the past literature has targeted this problem, which can be clearly seen in Table \ref{tab:taxonomy}. A good feature refinement module should be able to refine the node and edge features by analyzing both spatial as well as statistical features present in the image in question. While the traditional techniques for relationship detection in such visual relationship tasks involved the usage of CRF \cite{sutton2012introduction}, it has since been replaced with more efficient techniques.
\vspace{5mm}

\begin{table}
  \captionsetup{labelfont=bf}
  \centering
  \begin{tabular}{|P{4cm} P{1.9cm} P{1.9cm} P{1.9cm} P{1.9cm}|}
    \toprule
        Methods & Efficient Graph Features Refinement & Efficient Graph Generation & Long-tailed Dataset Distribution & Efficient Loss Function Definition \\
        \midrule
        \midrule
        Deep Relational Networks \cite{dai2017detecting} & \checkmark &  &  &  \\
        \midrule
        Iterative Message Passing \cite{xu2017scene} & \checkmark &  &  &  \\
        \midrule
        MSDN \cite{li2017scene} & \checkmark &  &  &  \\
        \midrule
        Factorizable Net \cite{li2018factorizable} &  & \checkmark &  &  \\
        \midrule
        Neural Motifs \cite{zellers2018neural} & \checkmark &  & \checkmark &  \\
        \midrule
        Graph R-CNN \cite{yang2018graph} & \checkmark & \checkmark &  &  \\
        \midrule
        Knowledge-embedded Routing Network \cite{chen2019knowledge} & \checkmark &  & \checkmark &  \\
        \midrule
        Graph Contrastive Losses \cite{zhang2019graphical} &  &  &  & \checkmark \\
        \midrule
        VCTree \cite{tang2019learning} & \checkmark &  & \checkmark &  \\
        \midrule
        External Knowledge and Image Reconstruction \cite{gu2019scene} & \checkmark &  & \checkmark &  \\
        \midrule
        Prediction with Limited Labels \cite{chen2019scene} &  &  & \checkmark &  \\
        \midrule
        Counterfactual Critic Multi-agent Training \cite{chen2019counterfactual} &  &  &  & \checkmark \\
        \midrule
        Visual Relationships as Functions \cite{dornadula2019visual} &  &  & \checkmark &  \\
        \midrule
        Unbiased SGG from Biased Training \cite{tang2020unbiased} &  &  & \checkmark &  \\
    \bottomrule
  \end{tabular}
  \newline\newline
  \caption{Taxonomy for the various Scene Graph Generation (SGG) methods. The tick in a particular column represents the problem that the method tries to target through its main contributions.}
  \label{tab:taxonomy}
\end{table}

\textbf{\emph{Methodology}} ~~ Deep Relational Networks \cite{dai2017detecting} and Iterative Message Passing \cite{xu2017scene} were two of the earliest techniques solving the SGG task by carefully taking care of the spatial as well as statistical features in a scene.

\emph{Deep Relational Networks} proposed in \cite{dai2017detecting} inferred the class labels by exploiting both spatial and statistical dependencies in the given scene. The spatial configuration between objects is responsible for seeing the proximity as well as relative spatial arrangement between the objects, while statistical dependency basically means the prevalence of certain relationships given prior information about the subject and object, like \emph{cat eat fish} is much more likely than \emph{fish eat cat}. Hence, the initial feature vectors for subject and object are taken to be the features obtained from bounding boxes of detected objects, while the initial relationship vectors are pre-encoded with prior spatial information by using features from the union of spatial masks of the subject and object. Now to refine these features adhering to the statistical dependencies, \emph{DR-Net} is used to obtain the class probabilities of the subject, object, and relationships by taking into account the feature vectors of all three simultaneously and refining on the output probabilities iteratively. The iterative process can be summarized in the equations stated below.
\begin{align}  \label{eq:dr_net}
    q'_s = \sigma ( W_a x_s + W_{sr} q_r + W_{so} q_o), \notag \\
    q'_r = \sigma ( W_r x_r + W_{rs} q_s + W_{ro} q_o), \notag \\
    q'_o = \sigma ( W_a x_o + W_{os} q_s + W_{or} q_r).
\end{align}
Here $q_s$, $q_r$, $q_o$ represent the probability distribution over the classes for subject, relationship and object features and $q'_s$, $q'_r$, $q'_o$ represent the updated probabilities. $x_s$, $x_r$ and $x_o$ represent the initial subject, relation and object feature vectors respectively. $\sigma$ denotes the softmax activation, while the weight vectors $W$ specified with various notations will remain constant for handling a particular kind of feature vector (e.g., $W_a$ for handling subject and object features, while $W_{os}$ for handling the relation between subject and object), as shown in the equation, and will be updated when training the network. As can be seen, the process resembles a special sort of RNN structure.

\begin{figure}[t!]
\captionsetup{labelfont=bf}
\centering
\raisebox{-.5\height}{\begin{subfigure}[b]{0.44\textwidth}
\includegraphics[width=2.5in]{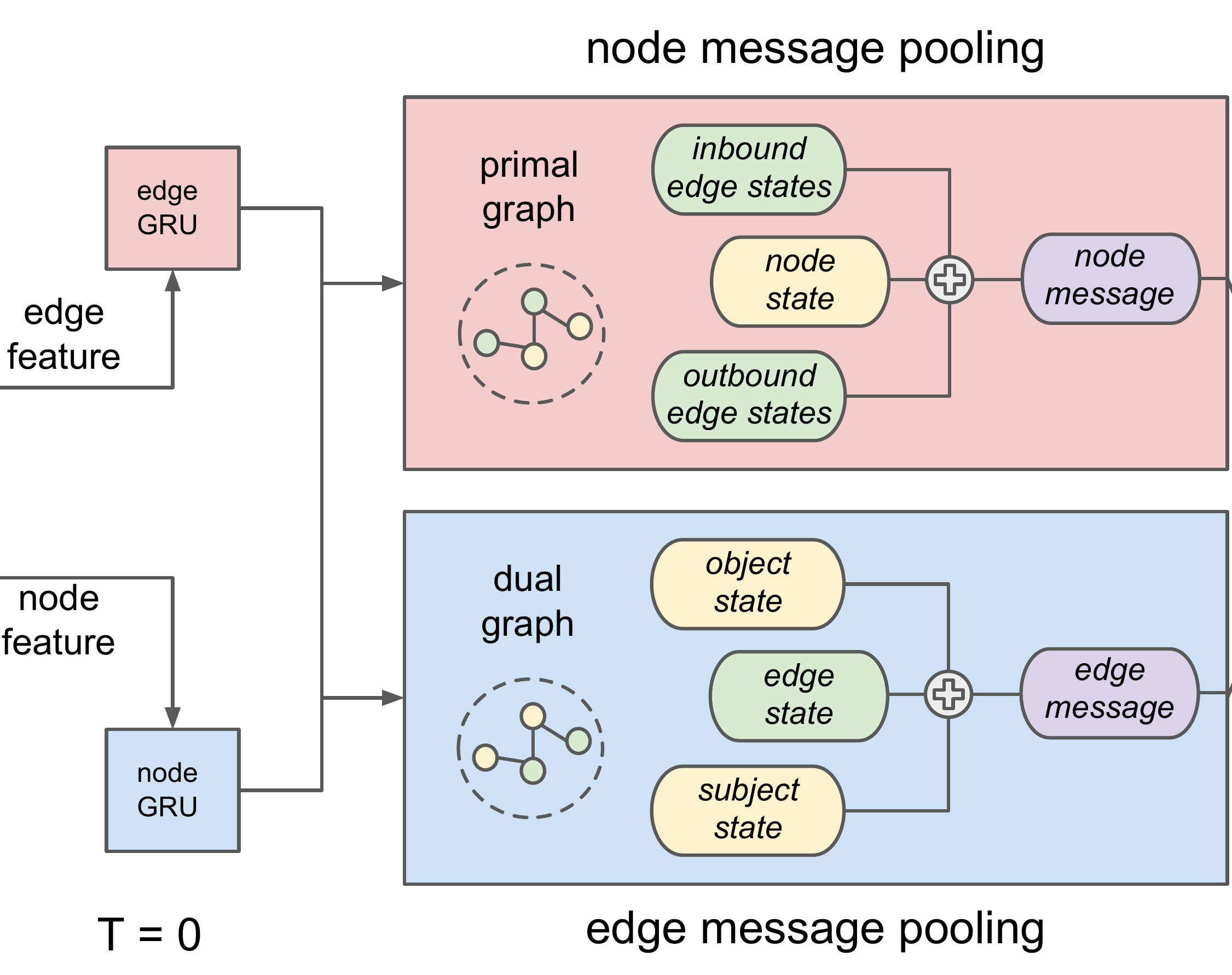}
\caption{}
\end{subfigure}}
\hspace{1cm}
\raisebox{-.5\height}{\begin{subfigure}[b]{0.44\textwidth}
\includegraphics[width=2.5in]{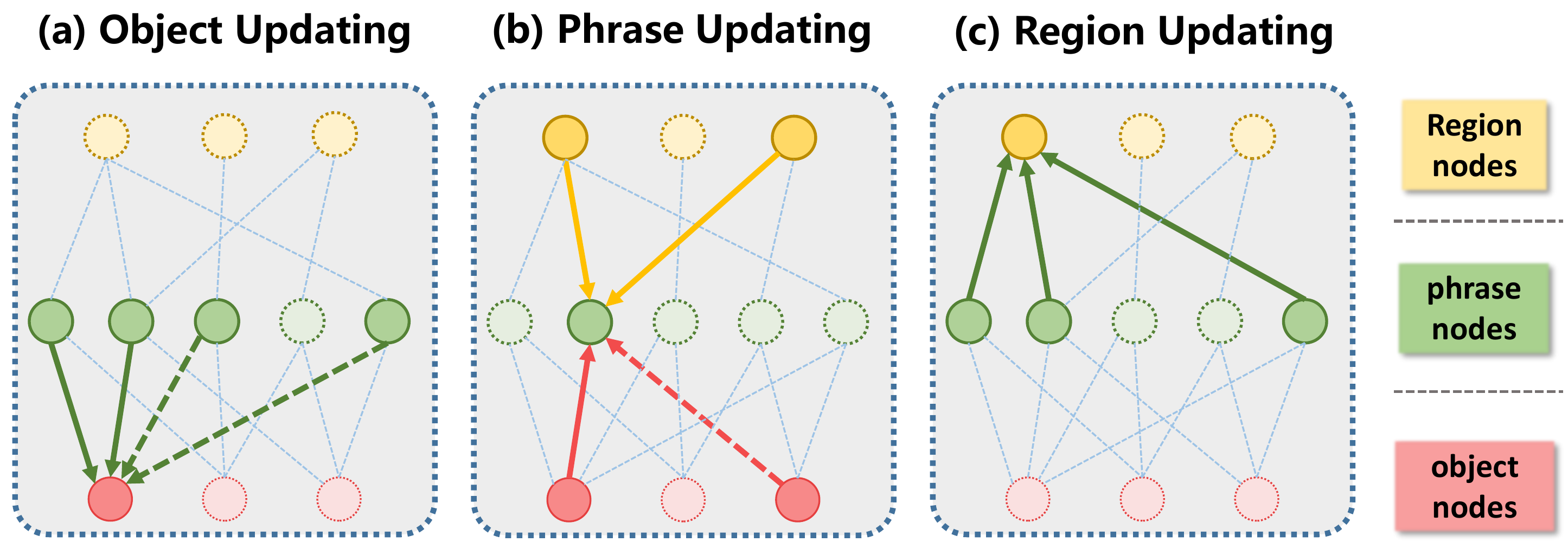}
\caption{}
\end{subfigure}}
\caption{The figures represent the feature refining steps for \cite{xu2017scene} and \cite{li2017scene}. The figure on the left (a) represents the Feature Refining Module proposed in \cite{xu2017scene}, with major concentration on the message pooling module for information interchange between the node and edge GRUs. The figure on the right (b) represents the feature refining process for \cite{li2017scene}, where the arrows represent the passing direction of information between the different kinds of features for updation.}
\label{fig:pipeline_iterative}
\end{figure}

Moving on similar lines of an updation mechanism by jointly optimizing node and edge features, \cite{xu2017scene} proposed an \emph{iterative message passing scheme}, much similar to the above but with node and edge features being represented by hidden states of GRUs, where common weights are shared for all nodes and separate common weights for all edges in the graph. This setup allows the model to pass messages among the GRU units following the scene graph topology. A \emph{message pooling module} is also designed for passing the messages between edges and nodes, since the set of edge GRUs and node GRUs form a bipartite graph. In the message pooling module, dual graphs are formed for edge and node GRUs. We have a node-centric primal graph, in which each node GRU gets messages from its inbound and outbound edge GRUs, while in the edge-centric dual graph, each edge GRU gets messages from its subject node GRU and object node GRU. To be more specific if $m_{i}$ and $m_{i\rightarrow j}$ are node and edge message that are to be passed for optimization then,

\begin{equation}  \notag
	m_i = 
    \sum_{j:i\rightarrow j}\sigma\left (v_{1}^T[h_i,h_{ i\rightarrow j }]\right)h_{i\rightarrow j} + \sum_{j:j\rightarrow i}\sigma\left (v_{2}^T[h_i,h_{ j\rightarrow i }]\right)h_{j\rightarrow i}
\end{equation}
\begin{equation}
	m_{i\rightarrow j} = 
	\sigma\left (w_{1}^T[h_i,h_{i\rightarrow j}]\right)h_i + 
	\sigma\left (w_{2}^T[h_j,h_{i\rightarrow j}]\right)h_j	 
\end{equation}
Where $h_i$ and $h_j$ are hidden state of subject and object respectively and $h_{i\rightarrow j}$, $h_{j\rightarrow i}$ are hidden states of outbound edge GRUs and inbound edge GRUs respectively for the $i$-th object. A visual representation of this pooling scheme can also be seen in Fig. \ref{fig:pipeline_iterative}(a).


Both the techniques exploited spatial as well as a statistical dependency for final prediction and jointly optimised the object and relation features rather than inferring them individually. Both the works also follow a similar iterative optimization approach. The success of these techniques signified the importance of joint optimization of node and edge features in an iterative format, and hence laid the groundwork for further work in this field. Building upon the strong correlation between the tasks of SGG, region captioning and object detection, \cite{li2017scene} proposed a \emph{Multi-level Scene Description Network (MSDN)} to perform all three tasks simultaneously and refine the features at different semantic levels to increase the abstraction power for a more accurate graph prediction. On the other hand, \cite{yang2018graph} proposed a variant of attentional GCN to propagate higher-order context throughout the graph and enable per-node edge attentions, allowing the model to learn to modulate information flow across unreliable and unlikely edges.

There are three kinds of region proposals in MSDN, namely object region proposals, phrase region proposals, and caption region proposals. Hence to update these features simultaneously, \cite{li2017scene} proposes making connections between object and phrase features and between phrase and caption features for a systematic flow of information between these three semantic levels, which can also be seen in Fig. \ref{fig:pipeline_iterative}(b). An updation scheme for phrase feature refinement can be seen as follows:

\begin{equation}
    {x}^{(p)}_{j, t+1} = ~ {x}^{(p)}_{j,t} + {F}^{(s\rightarrow p)}\left(\tilde{{x}}^{(s\rightarrow p)}_{j}\right) + {F}^{(o\rightarrow p)}\left(\tilde{{x}}^{(o\rightarrow p)}_{j} \right) + {F}^{(r\rightarrow p)}\left(\tilde{{x}}^{(r\rightarrow p)}_{j} \right)
\end{equation}
where ${x}^{(p)}_{j, t}$ represents the phrase features at time step $t$. $\tilde{{x}}^{(s\rightarrow p)}_{j}$ and $\tilde{{x}}^{(o\rightarrow p)}_{j}$ denote the merged features from subject and object respectively in the \emph{subject-predicate-object} triplet for the $j$-th phrase node, and $\tilde{{x}}^{(r\rightarrow p)}_{j}$ denotes the features merged from its connected caption nodes. Similar updation equations are defined for object and caption region proposals taking into account the connection dependencies between the semantic levels as also symbolized by Fig. \ref{fig:pipeline_iterative}(b).

\emph{MotifNet} \cite{zellers2018neural} builds on the hypothesis that the classes of the object pair are highly indicative of the relationship between them, while the reverse is not true. This is an example of the problem of \emph{Bias in Dataset}, which will be explored further in section \ref{class_long_tail}. Hence rather than following the traditional paradigm of bidirectional information propagation between object and relationships, they predict graph elements by staging bounding box predictions, object classifications, and relationships such that the global context encoding of all previous stages establishes a rich context for predicting subsequent stages. Hence the network decomposes the probability of a graph $G$ (made up of a set of bounding regions $B$, object labels $O$, and labeled relations $R$) into three factors:
\begin{equation}
Pr(G \mid I) = Pr(B \mid I) \ Pr(O \mid B, I) \ Pr(R \mid B, O, I)
\end{equation}
This kind of strong dependence assumption helps the model to capture the global context better. To model this, they have used BiLSTM for object context encoding and LSTMs for object context decoding. The obtained object classes are passed to a BiLSTM which encodes relationship context and predicts the predicates.

\begin{figure}[t!]
\captionsetup{labelfont=bf}
    \includegraphics[width=14cm]{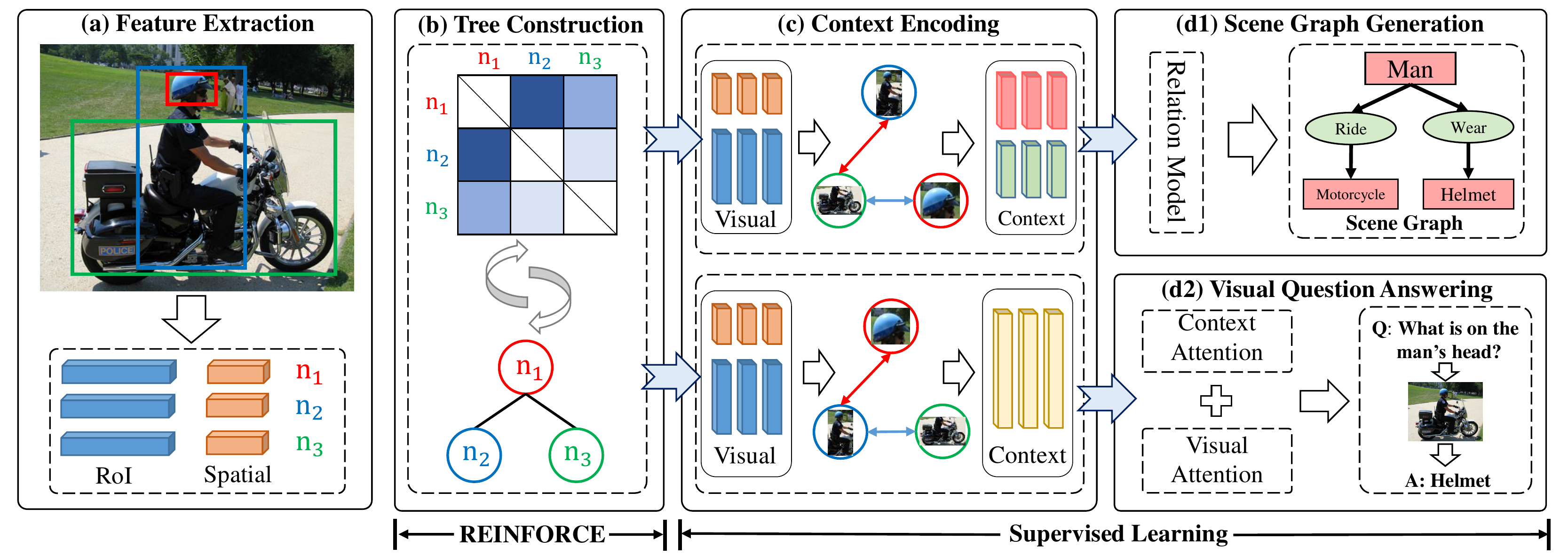}
\centering
\caption{The framework of the VCTree model \cite{tang2019learning}. Visual features are extracted from proposals, and a dynamic VCTree is constructed using a learnable score matrix. The tree structure is used to encode the object-level visual context, which will be decoded for each specific end-task. In context encoding (c), the right branches (blue) indicate parallel contexts, and left ones (red) indicate hierarchical contexts. Parameters in stages (c) \& (d) are trained by supervised learning, while those in stage (b) use REINFORCE with a self-critic baseline.} 
\label{fig:pipeline_vctree}
\end{figure}

All the techniques we have listed till now used LSTM or GCNs to refine the features, and while this kind of representation does seem sensible and has been great so far, one of the underlying problems with feature representation in this format is the lack of discriminative power between hierarchical relations in these. To solve this problem, \cite{tang2019learning} proposed a tree structure encoding for the node and edge features for both the SGG and VQA task, having it explicitly encode any hierarchical and parallel relations between objects in its structure. To construct the tree, a score matrix $S$ is first calculated to approximate the validity between object-pairs, with $S_{ij} = \sigma (MLP( x_i, x_j))$ where $x_i$ and $x_j$ are visual features of the object pair, after which Prim's algorithm is applied to obtain a Minimum Spanning Tree (MST). Once the multi-branch tree is constructed, it is converted into an equivalent binary tree \emph{VCTree}, an equivalent left-child and right-sibling tree (as can also be seen in Fig. \ref{fig:pipeline_vctree}(c)), which is done for discrimination between the hierarchical and parallel relations. Now, for encoding the context and refining these features, BiTreeLSTM \cite{tai2015improved}, is used in the following way:
\begin{equation}
D = \textnormal{BiTreeLSTM}(\{{z}_i\}_{i=1,2,...,n}),
\label{eq:bitreeLSTM}
\end{equation}
where ${z}_i$ is the input node feature, and $D=[{d}_1,{d}_2,...,{d}_n]$ is the encoded object-level visual context. Each ${d}_i=[\vec{{h}}_i;\cev{{h}}_i]$ is a concatenation of the hidden states from both TreeLSTM~\cite{tai2015improved} directions:
\begin{align}
&\vec{{h}}_i = \textnormal{TreeLSTM}({z}_i,\vec{{h}}_{p}), \label{eq:3}\\
&\cev{{h}}_i = \textnormal{TreeLSTM}({z}_i,[\cev{{h}}_{l};\cev{{h}}_{r}]),
\label{eq:treeLSTM}
\end{align}
where  $\vec{} ~~ \textnormal{and} ~~ \cev{}$ denote the top-down and bottom-up directions, and $p,l,r$ denote parent, left child, and right child of node $i$. For Object Context Encoding,  $z_i$ in Eq. \ref{eq:bitreeLSTM} is set to $[{x}_i;{W}_1 {\hat{c}}_i]$, a concatenation of object visual features and embedded class probabilities, where ${W}_1$ is the embedding matrix for the label distribution ${\hat{c}}_i$. The Relation Context Encoding is done by putting $z_i$ to be $d_i$ (features encoding object context) that has just been obtained using object context encoding step. Once the embeddings are created, decoding takes place, with the object class of a node dependant on its parent. For relationship extraction, three pairwise features are extracted for each object pair, namely context feature, bounding box feature, and RoIAlign feature, and the final predicate classification is done by combining all of them. Since the score matrix $S$ is not fully differentiable w.r.t the end task loss, a \emph{hybrid learning} strategy is followed, which combines policy gradient based reinforcement learning for the parameters $\theta$ of $S$ in the tree construction and supervised learning for the rest of the parameters.

\begin{figure}[b!]
\captionsetup{labelfont=bf}
    \includegraphics[width=14cm]{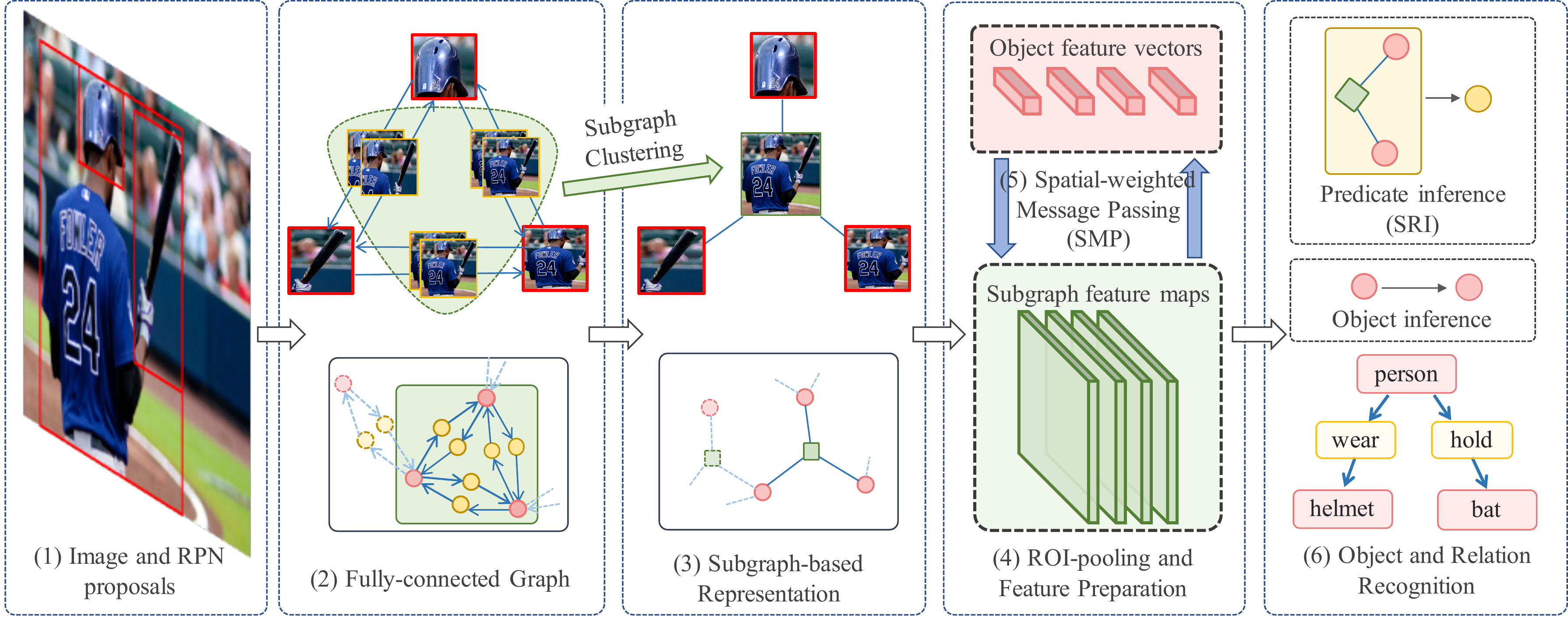}
\centering
\caption{Pipeline of \emph{Factorizable-Net} \cite{li2018factorizable}. (1) RPN is used for object region proposals, which shares the base CNN with other parts. (2) Given the region proposal, objects are grouped into pairs to build up a fully-connected graph, where each pair of objects are connected with two directed edges. (3) Edges that refer to similar phrase regions are merged into subgraphs, and a more concise connection graph is generated. (4) ROI-Pooling is employed to obtain the corresponding features~(2-D feature maps for subgraph and feature vectors for objects). (5) Messages are passed between subgraph and object features along with the factorized connection graph for feature refinement. (6) Objects are predicted from the object features and predicates are inferred based on the object features as well as the subgraph features. Green, red and yellow items refer to the subgraph, object and predicate respectively.}
\label{fig:pipeline_factorizable}
\end{figure}

\subsection{Efficient Graph Generation Task}
\label{class_graph_generation}
\textbf{\emph{Motivation}} ~~ While a good refinement module is of paramount importance in the SGG task, the creation of an efficient graph structure serves high importance. For the initialization of features, the most basic approach is to assume the creation of a fully connected graph once the object proposals are found, which means with $n$ number of detected objects, there will be $n(n-1)$ candidate relations. As is evident, the number of relation proposals present in a scene will quickly overshoot as the number of objects increases even slightly. Also, since not all objects will always have any significant relationship between them, having so many relationship proposals is redundant for model performance. We will look at some of the past work specifically trying to target this problem.
\vspace{5mm}

\begin{figure}[t!]
\captionsetup{labelfont=bf}
    \includegraphics[width=14cm]{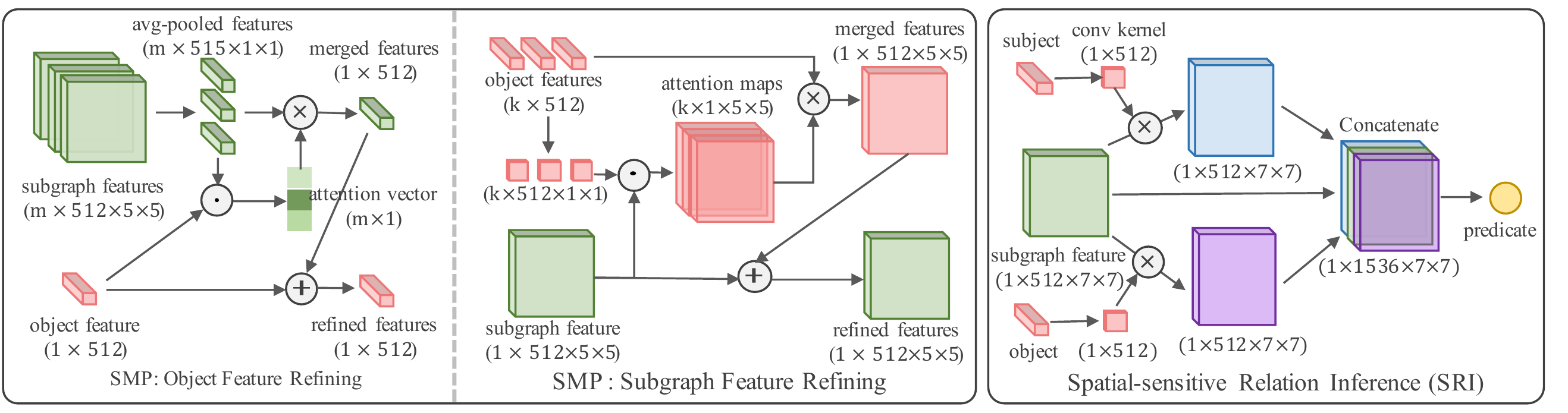}
\centering
\caption{Left: SMP structure for object/subgraph feature refining. Right: SRI Module for predicate recognition. Green, red and yellow refer to the subgraphs, objects, and predicates respectively. $\odot$ denotes the dot product, while $\oplus$ and $\otimes$ denote the element-wise sum and product, respectively.}
\label{fig:factorizable_modules}
\end{figure}

\textbf{\emph{Methodology}} ~~ One of the naive ways to solve the problem can be to remove some relationship edges randomly, which can undoubtedly solve the excessive number of relationships problem. However this can lead to a decrease in accuracy because of the removal of some important relationships due to its random nature. Hence, \emph{Factorizable Net} \cite{li2018factorizable} was proposed, and as the name suggests, it outlined a technique to break the main graph into several subgraphs based on some common features (see Fig. \ref{fig:pipeline_factorizable}). Rather than having a relationship proposal between every object, a subgraph is proposed as a relationship feature vector for a group of objects. For the creation of subgraphs, union box for two objects is taken and this box is given a confidence score as the product of the scores of two object proposals. Non-Max Suppression \cite{girshick2015fast} is then applied to get the representative box and a merged subgraph is formed, containing a unified feature representation for a number of objects. Furthermore, the edge features are represented by 2-D maps, while the object features are still  represented as 1-D vectors, prompting \cite{li2018factorizable} to introduce a novel \emph{Spatial-weighted Message Passing} (SMP) structure for message passing and also a \emph{Spatial-sensitive Relation Inference} (SRI) module to infer the predicate in the form of a 1D vector. The functioning of both of these can be seen in Fig. \ref{fig:factorizable_modules}.

While the above technique focuses on reducing the number of relations by not ignoring any relationship, \emph{Graph RCNN} \cite{yang2018graph} proposes a novel Relation Proposal Network (RePN) which efficiently computes relatedness scores between object pairs and prunes out unlikely relations. For this, given the initial object classification distributions $P^o$, a relatedness score $s_{ij} = f(p^o_i, p^o_j)$ is calculated where $f(. , .)$ is a learned relatedness function represented as:
\begin{equation}
    f(p^o_i, p^o_j) = \langle \Phi (p^o_i), \Psi (p^o_j) \rangle, i \neq j
\end{equation}
where $\Phi (.)$ and $\Psi (.)$ are projection functions for subjects and objects in a relationship respectively, and MLPs with the same structure are used to represent these projection functions. From, this we can select top $K$ relationships, and then Non-max suppression is used to filter out objects pairs with significant overlap with each other. Here the overlap between object pairs $(u, v)$ and $(p, q)$ is calculated as:
\begin{equation}
\small
IoU(\{u, v\}, \{p, q\}) = \frac{I({r}^o_u, {r}^o_p) + I({r}^o_v, {r}^o_q)}{U({r}^o_u, {r}^o_p) + U({r}^o_v, {r}^o_q)}
\end{equation}
where operator $I$ computes the intersection area between two boxes and $U$ the union area.

Once a sparse graph is obtained, attentional GCNs are used to refine the features, the propagation equations for which are represented as:
\begin{equation}
{z}_{i}^o = \sigma ( 
\overbrace{W^{\mathtt{skip}} Z^o {\alpha}^{\mathtt{skip}}}^{\substack{\text{Message from}\\ \text{Other Objects}}} + 
\overbrace{W^{sr} Z^r{\alpha}^{sr} + W^{or} Z^r{\alpha}^{or}}^{\substack{\text{Messages from}\\ \text{Neighboring Relationships}}}) 
\end{equation}

\begin{equation}
{z}_{i}^r = \sigma ({z}_{i}^r + \underbrace{W^{rs} Z^{o} {\alpha}^{rs} +  W^{ro} Z^{o} {\alpha}^{ro}}_{\text{Messages from Neighboring Objects}}).
\end{equation}

where $Z^o$ and $Z^r$ are object and relationship features, and $\alpha_i \in [0, 1]^n$ with entry 0 for nodes not neighboring $i$ and $\alpha_{ii} = 1$. Also, $W$ is the weight vector that is different for skip connections and for connecting two kinds of features, with $\sigma$ being an activation function.

\subsection{Long-tailed Dataset Distribution}
\label{class_long_tail}

\begin{figure}[b]
\captionsetup{labelfont=bf}
    \includegraphics[width=14cm]{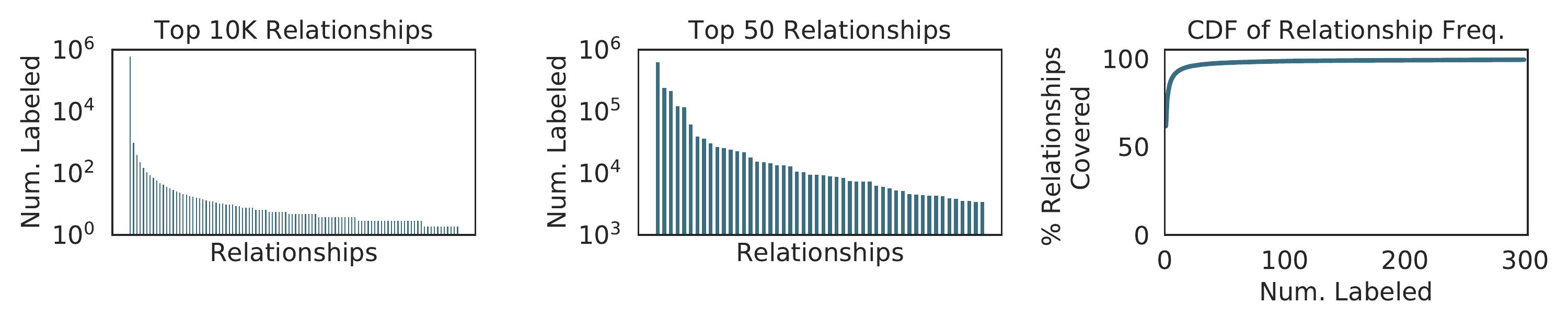}
\centering
\caption{Visual Relationships have a long tail (left) of infrequent relationships. Current models only focus on the top 50 relationships (middle) in the Visual Genome dataset, which all have thousands of labeled instances. This ignores more than $98\%$ of the relationships with few labeled instances (right, top/table).}
\label{fig:long_tail}
\end{figure}

\textbf{\emph{Motivation}} ~~ While the presence of large-scale datasets such as Visual Genome \cite{krishna2017visual} has undoubtedly been a huge turning point for the SGG task, it has come with some of its own problems, one of the major ones being the presence of a  \emph{long-tailed distribution} in the dataset. This refers to the presence of an uneven number of relationship instances in the dataset, with some of the simpler relations having many more instances (head) than more complex (and more informative) ones (tail). Consequently, while the network may be able to get good accuracy on paper, it might still fail to instigate diversity in its predictions, leading to performance deterioration in various downstream tasks such as VQA, image captioning, etc. The long-tail can be due to various reasons, one of the basic ones being the presence of certain \emph{bias} in human annotators towards simpler relationships such as \texttt{near}, \texttt{on}, \texttt{has}, etc instead of more complex ones such as \texttt{standing on}, \texttt{sitting on}, etc. A representation of this long-tail problem in the Visual Genome dataset can be seen in Fig. \ref{fig:long_tail}.
\vspace{5mm}

\textbf{\emph{Methodology}} ~~ The suggestion of the presence of such a bias in the Visual Genome dataset was first presented in \cite{zellers2018neural}. While their work does not directly solve the problem of bias presence, and rather focuses on designing a refining scheme to utilize this as we have already seen in Section \ref{class_feature_refinement}, it did report a very shocking baselines: \emph{given object detections, predict the most frequent relation between object pairs with the given labels, as seen in the training set}. While the baseline seems pretty simple, it did prove to be pretty powerful by improving on previous state-of-the-art by an average of $3.6\%$ relative improvement across evaluation settings. This clearly indicates a certain bias in the dataset towards some particular relation types, with a class imbalance being pretty apparent. With their technique, while they were able to beat their own baseline by capturing a better global context, it did not explicitly target these \emph{tail relationship classes}.

\begin{figure}[t!]
\captionsetup{labelfont=bf}
    \includegraphics[width=10cm]{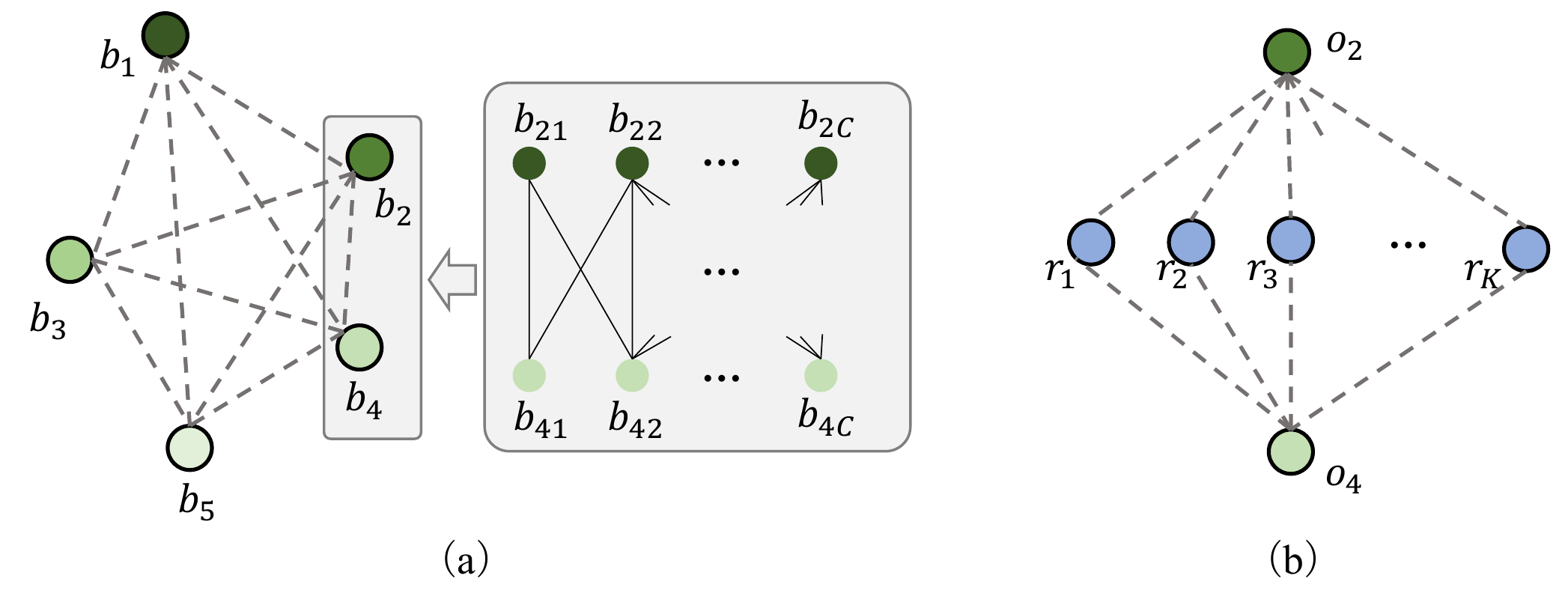}
\centering
\caption{Linking of various nodes in \cite{chen2019knowledge}. (a) A graph correlating the detected regions appearing in an image. (b) A graph correlating given object pair $o_i$ and $o_j$ with all the relationships.}
\label{fig:component_knowledge_embedding}
\end{figure}

\emph{Knowledge-embedded Routing Networks} \cite{chen2019knowledge} also tries to target the problem of uneven relationship distribution by using prior knowledge of statistical correlations between object pairs and their relationships. While \emph{MotifNet} \cite{zellers2018neural} implicitly tries to find statistical correlations, this was the first work to explicitly model these data statistics in its pipeline and regularize the semantic space of relationship prediction given a target object pair, hence taking the first step towards solving this problem. To do so, a $C \times C$ matrix $M_c$ is computed containing probabilities of occurrence of one class given another, where $c$ is the number of object classes. To explicitly represent this in the model, all object features $b_i$ are duplicated $C$ times to obtain $C$ nodes ${b_{i1}, b_{i2},..., b_{iC}}$ with node $b_{ic}$ denoting the correlation of region $b_i$ with category $C$. With the same process repeated for every node, $m_{cc'}$ is used to correlate node $b_{jc'}$ to $b_{ic}$, and in this way, $M_c$ can be used for calculating the correlation between every node pair. Now to obtain one object prediction per node, an updation scheme inspired by Graph Gated Neural Network \cite{li2015gated} is used for feature refining and final object class prediction. Once object classes are predicted, relationship classification takes place which again incorporates the statistical occurrence of a relationship given an object pair by finding the probabilities of all possible relationships given a subject of a category $c$ and an object of a category $c'$, which are denoted as ${m_{cc'1}, m_{cc'2},.., m_{cc'K}}$, where $K$ is the number of relationship types. As done previously, between any object pair, these probabilities are used to assign various relationship possibilities at the same time, and once again using the same architecture of gated GNN, the features are updated to obtain the final prediction. The connections between various nodes, in the case of object and relationship detection, can also be viewed in Fig. \ref{fig:component_knowledge_embedding}.

Another approach proposed in \cite{gu2019scene} is the usage of an external \emph{knowledge bank} to incorporate common sense into the network, for which they have used ConceptNet \cite{speer2013conceptnet}, as well as recontructing the image from the constructed scene graph to solve the problem of noisy labels in the dataset. To do this, object confidence scores are used to predict object label $a_i$, which is used to extract features from external knowledge bases:
\begin{align}
{a}_i&\overset{\text{retrieve}}{\longrightarrow} \langle {a}_i,{a}_{i, j}^r, {a}_{j}^o, w_{i,j}\rangle, j\in [0,K-1]\label{eq:concept_net_triplet}
\end{align}
where $a^r_{i,j}$, $a^o_{j}$ and $w_{i,j}$ are the top-$K$ corresponding relationships, the object entity and the  weight indicating how common a triplet $\langle {a}_i,{a}_{i,j}^r, {a}_{j}^o\rangle$ is. The triplets are then mapped to a sequence of words and transformed into a continuous vector space using a GRU to get a set of fact embeddings $F$. To extract the most relevant facts from this for feature refinement of an input object vector, an improved version of DMN \cite{xiong2016dynamic} is used. Firstly, an episode state is calculated using attention between the fact embeddings and the input object vector, after which the memory is updated using the current episode and previous memory state. Lastly, the final episodic memory is used to refine the object features, and the combined object and edge features are refined using the scheme in \cite{li2018factorizable}. Image reconstruction is also done for additional regularization, hence using a combination of scene graph level and image level supervision in this technique.

While the above models worked on using statistical information to get the desired result, few-shot learning and semi-supervised techniques have also been used in various literature works to treat the dataset imbalance. \cite{chen2019scene} and \cite{dornadula2019visual} enlist some ways to solve the long-tail problem in the SGG task by taking inspiration from the recent advances in semi-supervised and few-shot learning techniques. \cite{chen2019scene} proposes a semi-supervised approach by using image-agnostic features such as object labels and relative spatial object locations from a very small set of labeled relationship instances to assign probabilistic labels to relationships in unlabelled images. For this, image-agnostic features are extracted from objects in the labeled set $D_p$, and the object proposals from unlabeled set $D_U$. To capture the image-agnostic rules that define a relationship, heuristics are generated over these features with the help of decision trees. Labels are predicted using these heuristics, producing a matrix $\Lambda \in \mathbb{R}^{J \times |D_U|}$ of predictions for the unlabeled relationships, where $J$ is the number of decision trees. A factor graph-based generative model \cite{ratner2016data,xiao2015learning,roth2013combining} learns the accuracies of each heuristic to combine their individual labels, and outputs a probabilistic label for each object pair. Finally, these probabilistic labels can be used to train a scene graph prediction model. Also, a noise-aware empirical risk minimizer is used instead of a normal cross-entropy loss to take into account the errors in the training annotations.
\begin{figure}[t!]
\captionsetup{labelfont=bf}
    \includegraphics[width=14cm]{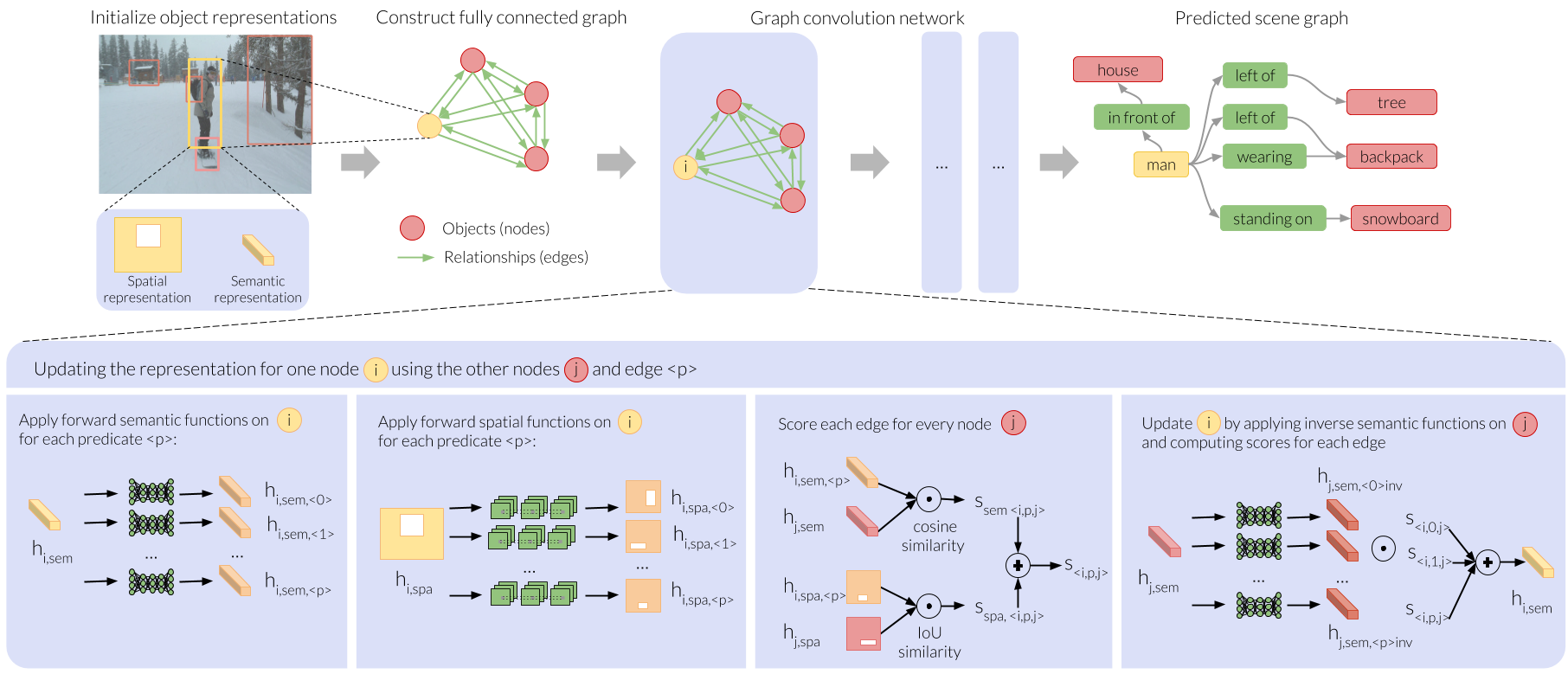}
\centering
\caption{Pipeline for SGG used in \cite{dornadula2019visual}. The formulation treats predicates as learned semantic and spatial functions, which are trained within a graph convolutional network. Objects are represented as semantic features and spatial attentions. The object representations form the nodes, and the predicate functions act as edges. Here the updation of one node (\texttt{person's} representation) is shown in one graph convolutional step.}
\label{fig:pipeline_few_shot}
\end{figure}

While the above technique used a model to label unlabeled data, \cite{dornadula2019visual} proposes a way to utilize the recent advances in few-shot learning for learning object representations from frequent categories and using these for few-shot prediction of rare classes. This is done by using a mechanism to create object representations that encode relationships afforded by the object. The idea is if a subject representation is transformed by a specific predicate, the resulting object representation should be close to other objects that afford similar relationships with the subject. For example, if we transform the subject, \texttt{person}, with the predicate \texttt{riding}, the resulting object representation should be close to the representations of objects that can be ridden and also at the same time be spatially below the subject. For such transformations, we deal with predicates as functions, dividing it into two individual functions: a forward function that transforms the \texttt{subject} representation into \texttt{object} and an inverse function that transforms the \texttt{object} representation back into \texttt{subject}. Each of these is further divided into two components: a spatial component ($f_{sem,p}$) that transforms attention over the image space and a semantic component ($f_{spa,p}$) that operates over the object features.

The general pipeline for the model in \cite{dornadula2019visual} can also be seen in Fig. \ref{fig:pipeline_few_shot}. During candidate graph construction, unlike previous works, a fully-connected graph is initialized, i.e., all objects are connected to all other objects by \emph{all predicate edges}. Hence for the convolution step, we will have message contributions for a subject node from all objects connected with all predicate edges. Also, rather than the traditional hidden representation of a node $h_i \in \mathcal{R}_D$, the hidden representation is a tuple of two entries: $h_i^t = (h_{i,sem}^t, h_{i,spa}^t)$ --- a semantic object feature $h_{i,sem}^t \in \mathcal{R}^D$ and a spatial attention map over the image $h_{i,spa} \in \mathcal{R}^{L\times L}$. Also, the semantic object feature $h_{i,sem}^t$ is updated while the spatial object feature $h_{i,spa}$ remains constant, where it is initially taken as an $L \times L$ mask with $1$ for the pixels within the object proposal and $0$ outside. However, during updation, without knowledge of which edge exists, a lot of noise will be added if every predicate is allowed to influence another node equally. To circumvent this issue, a score ($s_p(h_i^t, h_j^t)$) is calculated for each predicate $p$, by taking the convex combination of the spatial and semantic embeddings present in the subject's hidden representation. A rough method for calculation of scores can also be seen in Fig. \ref{fig:pipeline_few_shot}. Hence for the updation message of a subject node $i$'s hidden representation, the contribution from a particular predicate edge $p$ connecting to an object node $j$ can be calculated as:
\begin{equation}
    M_{sem}(h_{i,sem}^t,h_{j,sem}^t,e_{ijp}) = 
    s_{p}^l(h_{i}^t,h_{j}^t)f_{sem,p^{-1}}^{-1}(h_{j,sem}^t)
\end{equation}
where $M_{sem}$ is a learned message function specifically for semantic information and $f_{p^{-1}}(.)$ represents the backward predicate function from \texttt{object} back to the \texttt{subject}. Using these, the hidden state ($h_i^t$) can be updated as shown in Fig. \ref{fig:pipeline_few_shot}. Now these hidden states can be used to predict the objects and each possible relationship $e_{ijp}$ is output as a relationship only if $s_{p}^T(h_i^T, h_j^T) * s^{-T}_{p^{-1}}(h_j^T, h_i^T) > \tau$, where $T$ is the total number of iterations in the model and $\tau$ is a threshold hyperparameter. Once the functions are trained, they can be used for few-shot cases easily.
\begin{figure}[b!]
\captionsetup{labelfont=bf}
\centering
\raisebox{-.5\height}{\begin{subfigure}[b]{0.45\textwidth}
\includegraphics[width=2.5in]{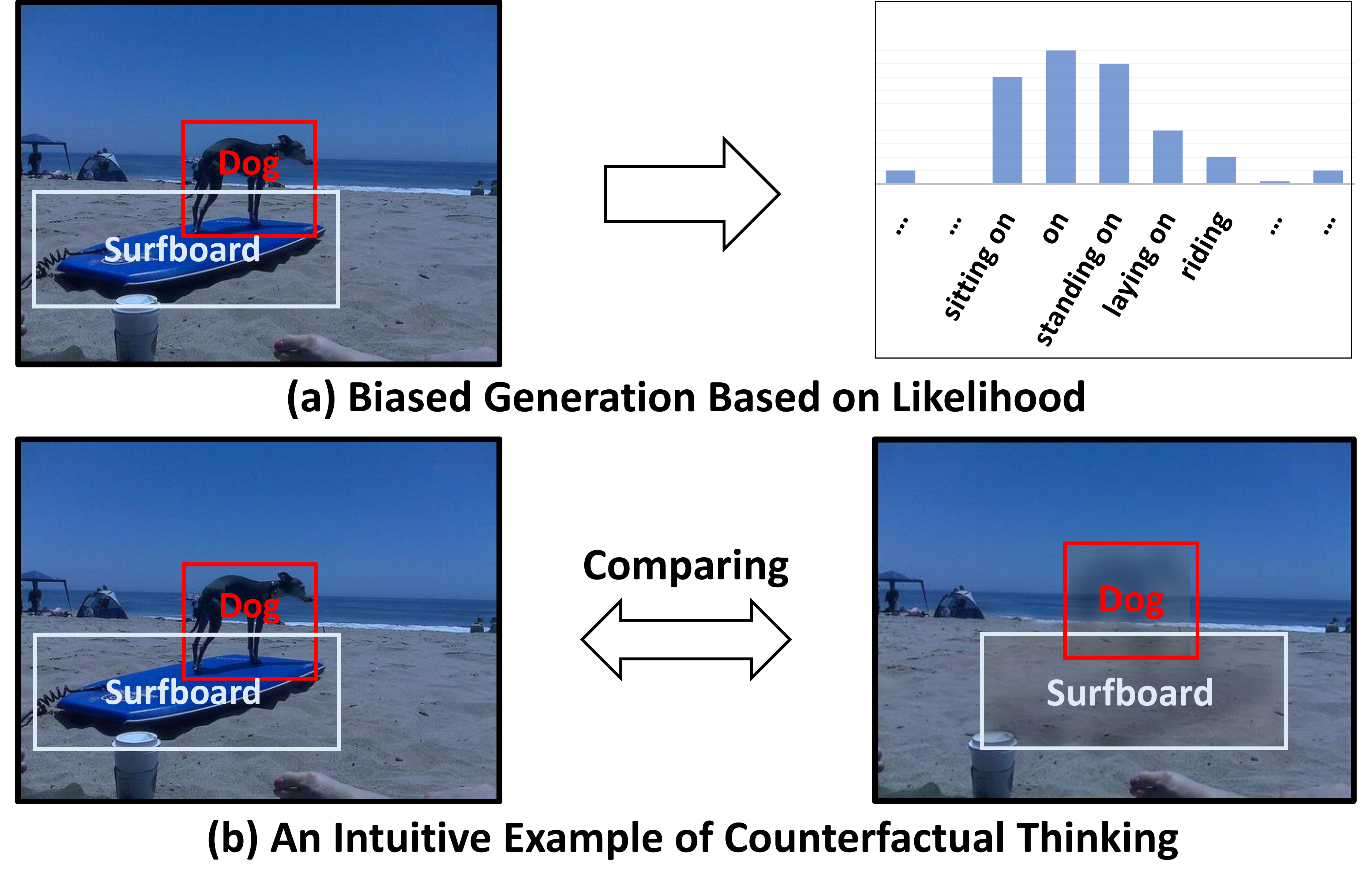}
\caption{\emph{(Top)} The biased generation that directly predicts labels from likelihood. \emph{(Bottom)} An intuitive example of the proposed Total Direct Effect (TDE), which calculates the difference between the real scene and the counterfactual one.}
\end{subfigure}}
\hspace{1cm}
\raisebox{-.5\height}{\begin{subfigure}[b]{0.45\textwidth}
\includegraphics[width=2.5in]{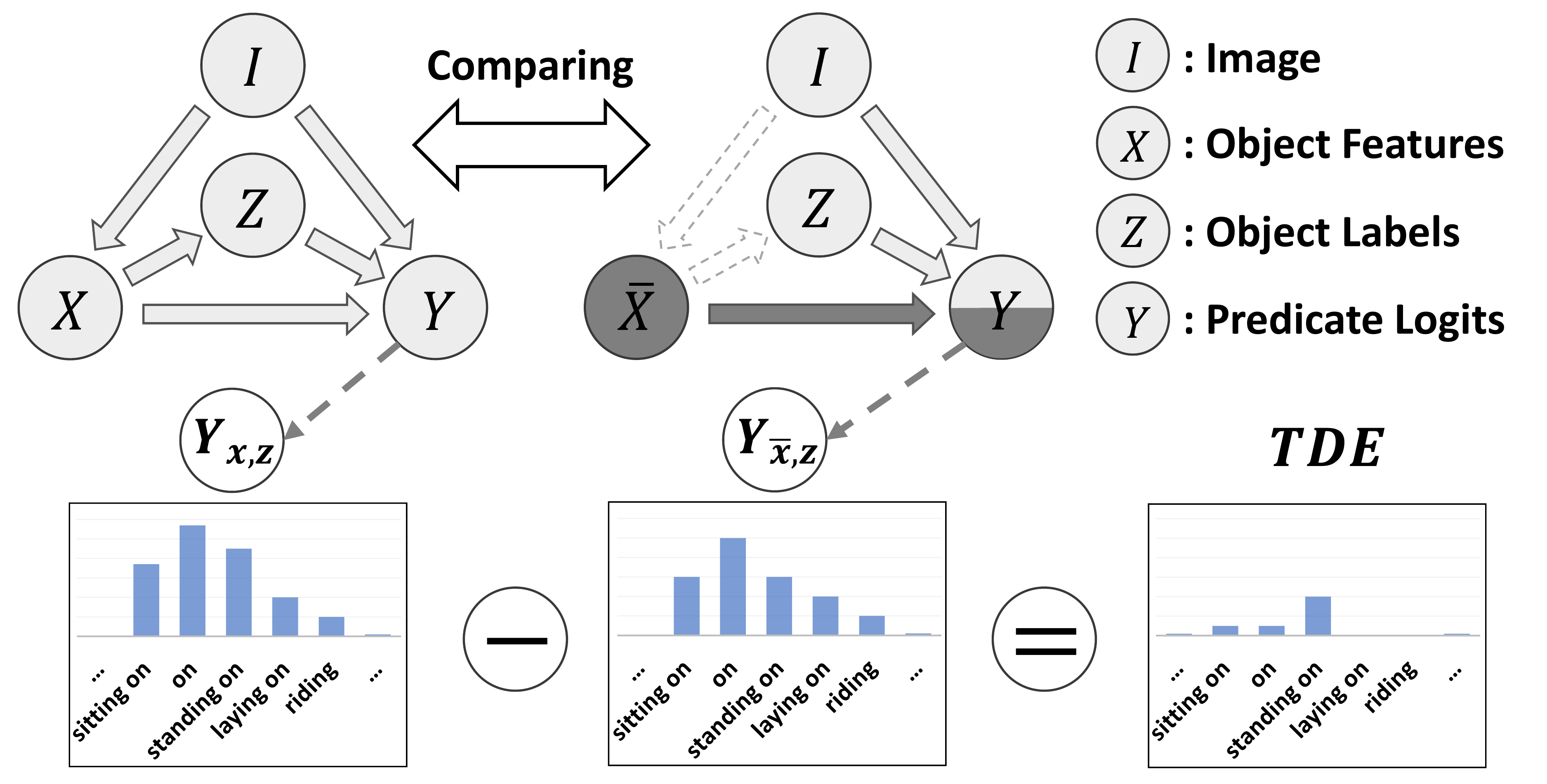}
\caption{An example of Total Direct Effect (TDE) calculation and corresponding operations on the causal graph, where $\bar{X}$ represents wiped-out $X$.}
\end{subfigure}}
\caption{\emph{(Left)} Represents the visual example of the framework proposed in \cite{tang2020unbiased}; \emph{(Right)} Represents the causal graphs of the normal and counterfactual scene.}
\label{fig:unbiased_from_biased}
\end{figure}

While all of the above techniques required training from scratch to solve the problem and essentially propose a complete model in themselves, \cite{tang2020unbiased} proposed a unique framework that can be used on top of any trained SGG model to deal with the \emph{bias} present in the dataset which eventually leads to the presence of a long-tail distribution in them. The framework, utilizing the concepts of \emph{causal inference}, can be attached on top of any model to make the model's relationship predictions less biased. A visual example of the framework proposed can also be seen in Fig. \ref{fig:unbiased_from_biased}. It proposes to empower machines with the ability of \emph{counterfactual causality} to pursue the "main effect" in unbiased prediction: '\emph{If I \textbf{had not} seen the content, would I still make the same prediction}'.

The novel unbiased SGG method proposed in \cite{tang2020unbiased} is based on the Total Direct Effect (TDE) analysis framework in causal inference \cite{pearl2001direct}. A visualization of the causal graph for the whole process can also be seen in Fig. \ref{fig:unbiased_from_biased}, which represents both the pipeline of biased training and that of unbiased training suggested in \cite{tang2020unbiased}. Hence to have an unbiased prediction by causal effects, we will intervene in the value of $X$ (object features) \emph{without changing any of the other features ($Z$ in this case)}, whether they are dependant on $X$ or not. The observed $X$ is denoted as $x$ while the intervened unseen value is $\bar{x}$, which is set to either the mean feature of the training set or zero vector. At last, the proposed unbiased prediction $y^{\dagger}_{e}$ is obtained by replacing the conventional one-time prediction with TDE, which essentially ``thinks'' twice: once for observational $Y_{x_{e}}(u)=y_{e}$, the other for imaginary $Y_{\bar{x},z_e}(u) = y_{e}(\bar{x}, z_{e})$. The unbiased logits of Y are therefore defined as follows:
\begin{equation}
    y^{\dagger}_{e} = y_{e} - y_{e}(\bar{x}, z_{e}).
\end{equation}
This gives logits free from context bias.

\subsection{Efficient Loss Function Definition}
\label{class_loss_func}
\textbf{\emph{Motivation}} ~~ In most of the techniques we have gone through till now for scene graph generation, the end task loss that has been used is a simple likelihood loss. While the same loss has been used for various tasks in Deep Learning literature, a more intricate loss designed just for relationship detection accounting for certain edge cases and certain problems specific to this task could certainly help to further improve model performance. Hence a small chunk of the past literature has focused on improving the overall accuracy by making some modifications in the end-task loss to target relationship detection problems specifically.
\vspace{5mm}

\textbf{\emph{Methodology}} ~~ Graphical \emph{Contrastive Losses} are suggested in \cite{zhang2019graphical} specifically for two kinds of problems, \emph{Entity Instance Confusion}, which occurs when a model confuses multiple instances of the same type of entity (e.g. multiple cups) and \emph{Proximal Relationship Ambiguity}, arising when multiple subject-predicate-object triplets occur in close proximity with the same predicate and the model struggles to infer the correct subject-object pairings (e.g. mis-pairing musicians and their instruments). The losses explicitly force the model to disambiguate related and unrelated instances through margin constraints specific to each type of confusion. The losses are defined over an affinity term $\Phi(s, o)$, interpreted as the probability that subject $s$ and object $o$ have some relation. For this, three separate contrastive losses are defined as follows:

\begin{enumerate}
    \item \emph{Class-Agnostic Loss} ~~ Used for contrasting positive/negative pairs regardless of their relation and adds contrastive supervision for generic cases. For a subject indexed by $i$ and an object indexed by $j$, the margins to maximize can be written as:
    \begin{equation}
    \begin{aligned}
    m_1^s(i)=\min_{j \in \mathcal{V}_i^+} \Phi(s_i,o_j^+)-\max_{k \in \mathcal{V}_i^-} \Phi(s_i,o_k^-) \\
    m_1^o(j)=\min_{i \in \mathcal{V}_j^+} \Phi(s_i^+,o_j)-\max_{k \in \mathcal{V}_j^-} \Phi(s_k^-,o_j)
    \label{eq:m_1}
    \end{aligned}
    \end{equation}
    where $\mathcal{V}_i^+$ and $\mathcal{V}_i^-$ represent sets of objects related to and not related to subject $s_i$, and a similar definition for $\mathcal{V}_j^+$, $\mathcal{V}_j^-$ and $o_j$. Finally, this loss can be defined as:
    \begin{align}
    \begin{split}
    L_1=&\frac{1}{N}\sum_{i=1}^{N}\max(0,\alpha_1-m_1^s(i)) \\ +&\frac{1}{N}\sum_{j=1}^{N}\max(0,\alpha_1-m_1^o(j))
    \end{split}
    \label{eq:class_agnostic}
    \end{align}
    where $N$ is the number of annotated entities and $\alpha_1$ is the margin threshold.
    
    \item \emph{Entity Class Aware Loss} ~~ Used for solving the \emph{Entity Instance Confusion} and can be taken as an extension of the above loss where class $c$ is defined when populating positive and negative sets $\mathcal{V}^+$ and $\mathcal{V}^-$. Hence the margins are defined in a similar way as Eq. \ref{eq:m_1}, where rather than $\mathcal{V}$, $\mathcal{V}^c$ is used to refer to class instances. Also, the definition of $L_2$ is pretty similar to Eq. \ref{eq:class_agnostic}, taking into account the margins specified here.
    
    \item \emph{Predicate Class Aware Loss} ~~ Used for solving the \emph{Proximal Relationship Ambiguity} problem and can be taken as an extension of the above two losses where relationship $e$ is defined when populating positive and negative sets $\mathcal{V}^+$ and $\mathcal{V}^-$. Hence the margins are defined in a similar way as Eq. \ref{eq:m_1}, where rather than $\mathcal{V}$, $\mathcal{V}^e$ is used to refer to a set of subject-object pairs where ground truth predicate between $s_i$ and $o_j$ is $e$. Also, the definition of $L_3$ is pretty similar to Eq. \ref{eq:class_agnostic}, taking into account the margins specified here.
\end{enumerate}

\begin{figure}[t]
\captionsetup{labelfont=bf}
\centering
\raisebox{-.5\height}{\begin{subfigure}[b]{0.45\textwidth}
\includegraphics[width=2.5in]{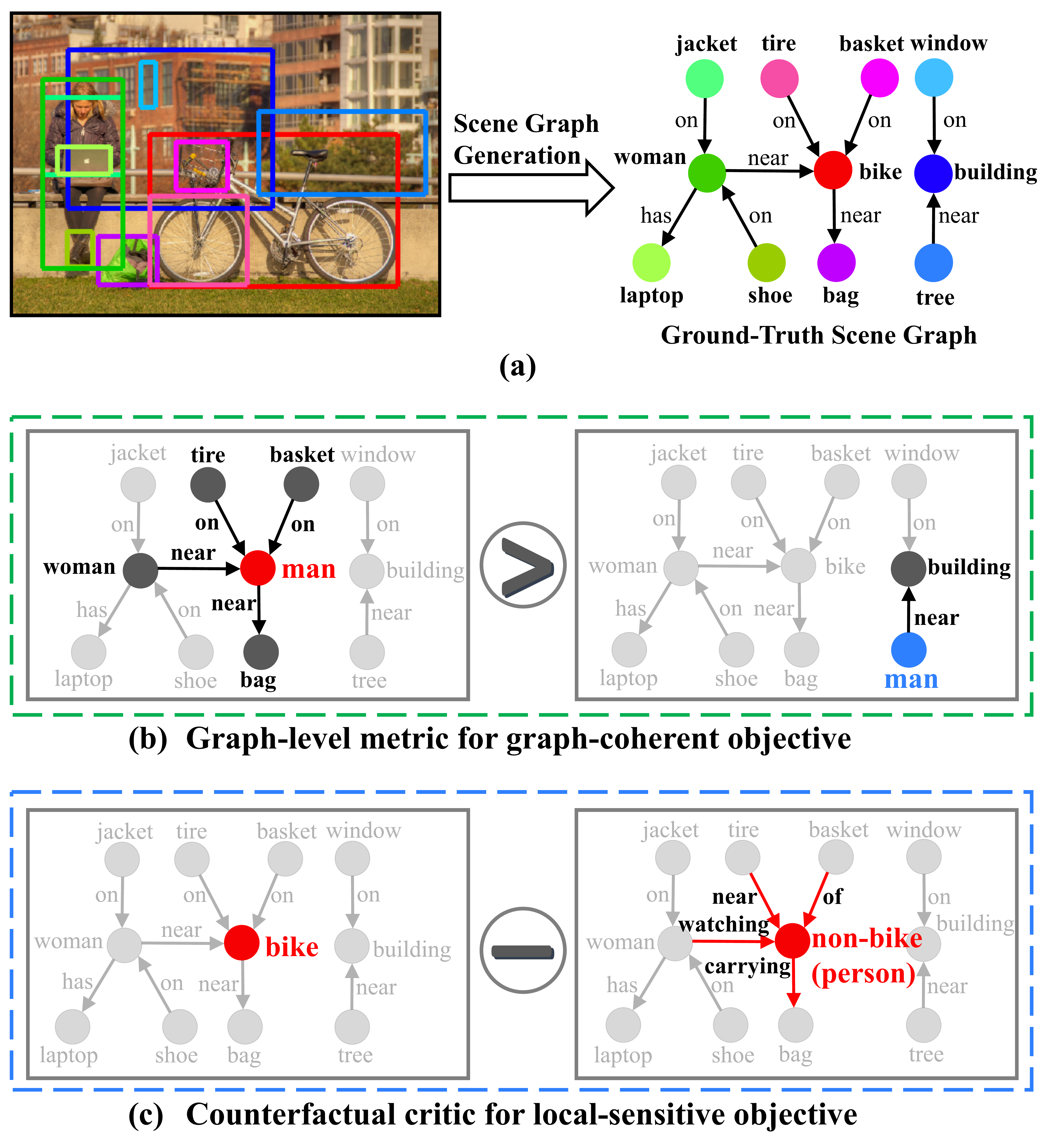}
\caption{(\emph{Top}) An input image and its ground-truth scene graph. (\emph{Middle}) For graph-coherent objective, a graph-level metric will penalize the red node \textbf{more than} $(>)$ the blue one, even though both are misclassified as \texttt{man}. (\emph{Bottom}) For local-sensitive objective, the individual reward for predicting the red node as \texttt{bike} can be identified by \textbf{excluding} $(-)$ the reward  from \texttt{non-bike} predictions.}
\end{subfigure}}
\hspace{1cm}
\raisebox{-.5\height}{\begin{subfigure}[b]{0.45\textwidth}
\includegraphics[width=2.5in]{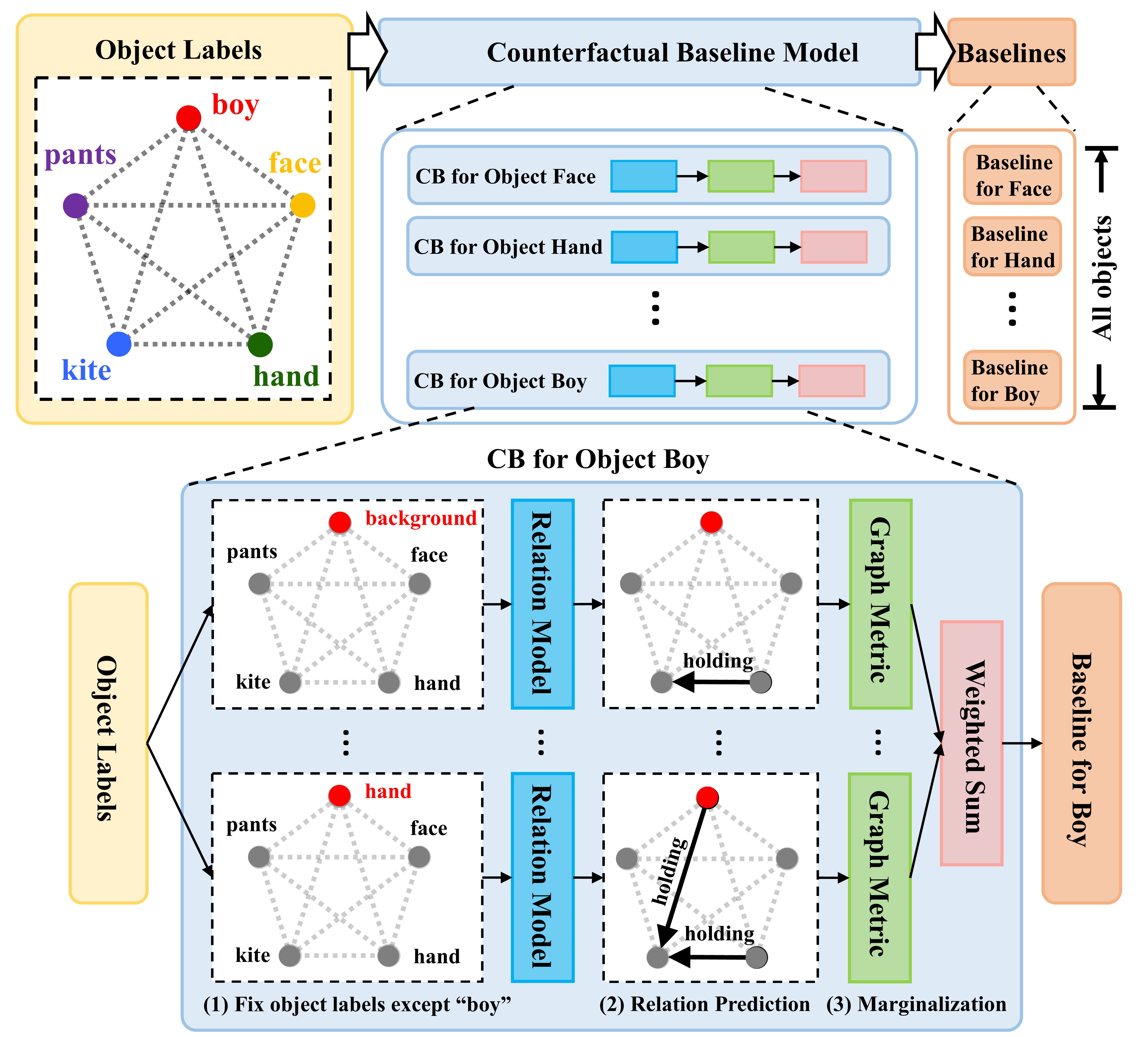}
\caption{The illustration of the counterfactual baseline (CB) model. For this given image, the model calculates CB for all agents. As shown in the bottom, for the CB for \textcolor{red}{\texttt{boy}}, we traverse to replace class label \texttt{boy} to all possible classes (\texttt{background} $,...,$ \texttt{hand}) and marginalize these rewards.}
\end{subfigure}}
\caption{(a) presents examples of the problem targeted in \cite{chen2019counterfactual}; (b) presents the procedure for calculating the counterfactual baseline CB.}
\label{counterfactual_multi_agents}
\end{figure}

A linear combination of all these losses, along with a standard cross-entropy makes up the final objective loss. In this way, \cite{zhang2019graphical} takes care of some of the edge cases having the problem types stated above. Another problem specified in \cite{chen2019counterfactual} is that of \emph{Graph Coherency} and \emph{Local-Sensitivity} in the objective. While \emph{Graph Coherency} means that the quality of the scene graph should be at graph-level and the detected objects and relationships should be contextually consistent, \emph{Local-Sensitivity} means that the training objective should be sensitive to the changes of a single node. Hence, \emph{Counterfactual critic Multi-Agent Training (CMAT)} \cite{chen2019counterfactual} is proposed to meet both the requirements. In this, a novel communicative \emph{multi-agent} model is designed, where objects are viewed as cooperative agents, and the action of each agent is to predict its object class labels. For the graph-coherent objective, the objective is defined as a graph-level reward (e.g., Recall@K \cite{lu2016visual} or SPICE \cite{anderson2016spice}) and policy gradients \cite{sutton2000policy} are used to optimize the non-differentiable objective, where the relationship model can be framed as a \emph{critic} and the object classification model serves as a policy network. For the local-sensitive objective, a counterfactual baseline is subtracted from the graph level reward by varying the target agent and fixing the others before feeding into the critic. An example of these two problems can also be seen in Fig. \ref{counterfactual_multi_agents}.

For a \emph{graph coherent} training objective, each detected object is modeled as an agent whose action is to predict the object class labels $v^t$. An LSTM is used to encode the \emph{history} of each agent, whose hidden state $h_i^t$ can be treated as a partially-observable environment state. Based on the policy gradient theorem, the gradient is given by 
\begin{align}
\begin{small}
\nabla_{\theta} J \approx \sum^n_{i=1} \nabla_{\theta} \log {p}^t_i (v^T_i|h^T_i; \theta) R(H^T, V^T),
\end{small}
\end{align}
where $R(H^T, V^T)$ is the real graph-level reward (Recall@K or SPICE).

To incorporate local sensitive signals to the objective, a counterfactual critic approach is used, where the individual contribution of an agent is represented by subtracting counterfactual baseline $ \text{CB}^i(H^T, V^T) = \sum {p}^T_i(\tilde{v}^T_i) R(H^T, (V^T_{-i}, \tilde{v}^T_i))$ from the global reward $R(H^T, V^T)$ to get the disentangled contribution of the action of an agent $i$,
\begin{align}
\begin{small}
    A^i(H^T, V^T) = R(H^T, V^T) - \text{CB}^i(H^T, V^T).
\end{small}
\end{align}
Here $A^i(H^T, V^T)$  can be considered as the \emph{advantage} and $\text{CB}^i(H^T, V^T)$ can be regarded as a \emph{baseline} in policy gradient methods. A visual representation of CB model can also be seen in Fig. \ref{counterfactual_multi_agents}. Hence the gradient becomes 
\begin{align} \label{eq:gradient}
\begin{small}
\nabla_{\theta} J \approx \sum^n_{i=1} \nabla_{\theta} \log {p}^t_i (v^T_i|h^T_i; \theta) A^i(H^T, V^T).
\end{small}
\end{align}
Finally, the cross-entropy losses are also incorporated in the overall gradient along with gradients in Eq. \ref{eq:gradient}.

%% file: datasets.tex
\section{Datasets}
\label{datasets}

Like other visual relationship detection tasks, Scene Graph Generation also requires a huge amount of annotated data. In this section, we will discuss the datasets commonly reported in SGG literature, majorly focusing on Visual Genome, a dataset specifically for the SGG task, while also discussing about some other datasets closely related to the task.

\subsection{Scene Graphs Dataset}
Scene Graphs Dataset \cite{johnson2015image} was the first dataset to introduce scene graphs grounded to images. It contains 5,000 images from the intersection of the YFCC100M \cite{thomee2016yfcc100m} and Microsoft COCO datasets \cite{lin2014microsoft}, allowing hierarchal development over these two. The full dataset of 5,000 images contains over 93,000 object instances, 110,000 instances of attributes, and 112,000 instances of relationships with an average of 2.3 predicates per object category. Since the predicates per object category are few, the dataset works fine for the image retrieval task but is insufficient for scene graph generation.

\subsection{Visual Relationship Dataset}
Though there existed other datasets that had annotations for relationships, they were either focused on object detection \cite{farhadi2011recognition} or image retrieval \cite{johnson2015image}. Also, existing datasets which focused on the task of relationship detection like the Visual Phrases \cite{farhadi2011recognition} dataset mainly covered relationships which were common. The Scene Graph Dataset \cite{johnson2015image} had 23,190 relationship types, but only 2.3 predicates/object which made relationship detection similar to the object detection task. To create a dataset benchmark especially for Visual Relationship detection, \cite{lu2016visual} introduced the Visual Relationship Dataset (VRD) which contains 5,000 images with 100 object categories and 70 predicate categories. It contains  37,993 relationships with 6,672 relationship types and an average of 24.25 predicates per object category.

\subsection{Visual Genome}

Visual Genome \cite{krishna2017visual} consists of 108,249 images from the intersection of 328,000 images of MS-COCO \cite{lin2014microsoft} and 100 million images of YFCC100M \cite{thomee2016yfcc100m} with three main motivations: 1) grounding visual objects to language, 2) providing a complete set of descriptions of a scene, 3) providing a formalized representation of an image. The dataset consists of 7 main components for each image: 

\begin{enumerate}

    \item \textbf{Region Descriptions}:
Previous datasets mainly focused on describing only the most important object in the image. Instead Visual Genome provides an average of 42 region descriptions per image to describe the image in a more comprehensive low-level manner.

    \item \textbf{Objects and Bounding Boxes}:
Previous datasets like MS-COCO \cite{lin2014microsoft} were unable to describe all the objects that appear in the real world since it only annotated 91 object categories. To fix this the Visual Genome dataset annotated over 17,000 semantic categories with around 21 objects per image with tight bounding boxes.

    \item \textbf{Attributes} : 
The dataset has on an average 16 object-specific attributes per image which allow for an easier classification and description of objects.

    \item \textbf{Relationships}:
The dataset has on an average 18 relationships per image, which help to represent connections between two objects in a graphical structure, ulitmately leading to a scene graph.

    \item \textbf{Question-Answer Pairs}:
The dataset contains 1.7 million QA's, with each image having at least one question of each of the 6 different what, where, how, when, who and why types.

    \item \textbf{Region Graphs}:
The dataset contains a graphical representation of each of the 42 regions per image, having the objects, attributes and relationships as nodes.

    \item \textbf{Scene Graph}:
A single scene graph is formed for an image by taking the union of all the region graphs of an image to get a graphical representation of the entire image.

\end{enumerate}

The dataset also employs Canonicalization, or WSD \cite{navigli2009word}, on all of its components to generalise them, so as to be connected with other datasets in research. 

As mentioned in \cite{chen2019scene}, the Visual Genome dataset has a long tail of infrequent relationships, and most models for scene graph generation that use it for training only focus on the top 50 relationships that have thousands of labels, ignoring almost 98 \% of the total relationship categories with limited labels. This problem was also discussed upon in Section \ref{methods}.

The dataset is used for a wide variety of tasks like scene graph generation, visual question answering, semantic image retrieval and relationship extraction.

\subsection{Action Genome}

Existing video datasets provide large amount of video clips with  action labels \cite{caba2015activitynet, carreira2019short, zhao2019hacs}, however they consider actions as monolithic events and do not study how objects and their relationships change during actions/activities. Taking inspiration from works in Cognitive Science, \cite{ji2019action} introduces Action Genome decomposing events into prototypical action-object units in the form of a temporal version of the Visual Genome's scene graphs called  spatio-temporal scene graphs. 

Based on the Charades dataset \cite{sigurdsson2016hollywood}, Action Genome provides scene graph labels for the components of each action in a video, providing annotations for 234,253 frames with 476,229 bounding boxes of 35 object classes and 1,715,568 instances of 25 relationship classes. It follows an action-oriented sampling strategy, providing more labels where more actions occur and divides human-object relationships into three categories: \emph{attention}, \emph{spatial}, \emph{contact}.

The authors also propose a method called Scene Graph Feature Banks (SGFB), which generates scene graphs for each frame followed by the construction of a feature bank from the scene graphs. The spatio-temporal information from the feature bank was shown to improve performances in action recognition on Charades, few-shot action recognition and also in the newly proposed task of spatio-temporal scene graph prediction which aims to predict a spatio-temporal scene graph from an input video stream.

%% file: applications.tex
\section{Applications}
\label{applications}

Having dealt with the problem of Scene Graph Generation, the obvious question that comes into the reader's mind is \emph{"why go through such trouble to create scene graphs to represent visual relationships in a scene?"}. Hence to make sense of all the work to improve the performance in the generation task, we are now going to delve into the various applications of scene graphs and how their rich representational power has been used to solve various downstream tasks such as VQA, Image Captioning, and many others.

\subsection{Semantic Image Retrieval}
Retrieving images by describing their contents is one of the most exciting applications of computer vision. An ideal system would allow people to search for images by specifying not only objects ("man" and "boat"), but also structured \emph{relationships} ("man on boat") and \emph{attributes} ("boat is white"). However as pointed out in \cite{johnson2015image}, bringing this level of semantic reasoning to real-world scenes involves two main challenges: (1) interactions between objects in a scene can be highly complex, going beyond simple pairwise relations, and (2) the assumption of a closed universe where all classes are known beforehand does not hold.

Hence to tackle these problems, \cite{johnson2015image} proposes the use of scene graphs for the task. Replacing textual queries (used for retrieval) with scene graphs allows the queries to represent the visual relationships in a scene in an explicit way rather than relying on unstructured text for the task. The paper introduced a novel Conditional Random Field (CRF) model for retrieving the image from a given scene graph, outperforming retrieval methods based on low-level visual features. For the CRF model formulation, let $G = (o, E)$ be a scene graph, $B$ be a set of bounding boxes in an image, and $\gamma$ be a grounding of the scene graph to the image. Each object $o \in O$ gives rise to a variable $\gamma_o$ in the CRF, where the domain of $\gamma_o$ is $B$. The distribution is modeled over possible groundings as:
\begin{equation}
    P(\gamma | G, B) = 
    \prod_{o \in O} P(\gamma_o | o)
    \prod_{(o, r, o') \in E} P(\gamma_o, \gamma_o' | o, r, o')
\end{equation}
The two product sequences represents the unary and binary potentials respectively, with the unary potential term modelling how well the appearance of the box $\gamma_o$ agrees with the known object classes and attributes of object $o$, and the binary potential term modelling how well the pair of bounding boxes $\gamma_o , \gamma_{o'}$ express the tuple $(o, r, o')$.

Further work by \cite{schuster2015generating} focuses on the fact that manual creation of scene graph is tough, hence working upon its automation by user encoding the relationships to parse the image description. The created scene graphs are used for image retrieval with the same CRF formulation as above. Here the scene graph creation acts as an intermediate step, rather than the input as in \cite{johnson2015image}. Even with textual query as the main input, they report almost the same performance for image retrieval when using human-constructed scene graphs. This result clearly implies a direct correlation between a more accurate image retrieval method and the Scene Graph Generation task.

\begin{figure}[t]
\captionsetup{labelfont=bf}
    \includegraphics[width=14cm]{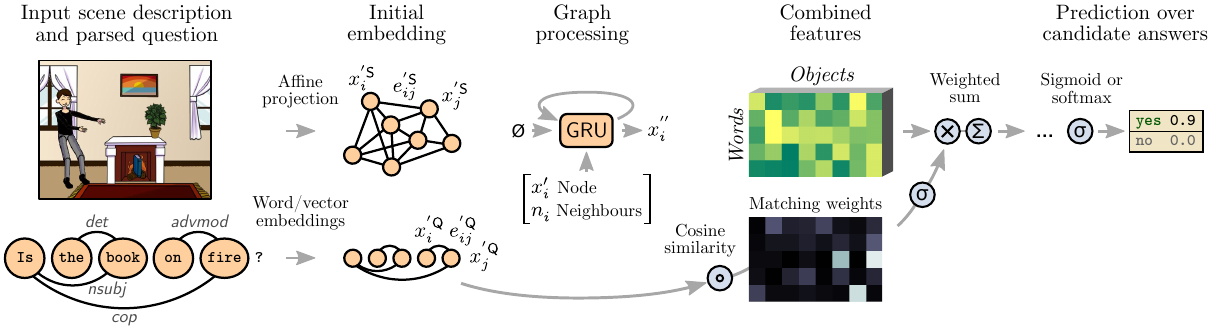}
\centering
\caption{A pipeline for \cite{teney2017graph} on using scene graphs for VQA. The input is provided as a description of the scene and a parsed question. A recurrent unit (GRU) is associated with each node of both the graphs that updates the representation of each node over multiple iterations. Features from all objects and all words are combined (concatenated) pairwise and they are weighted with a
form of attention. This effectively matches elements between the question and the scene. The weighted sum of features is passed through a final classifier that predicts scores over a fixed set of candidate answers.} 
\label{fig:vqa_pipeline}
\end{figure}

\subsection{Visual Question Answering}
With the rich representational power of scene graphs and the dense \& explicit relationships between various nodes (objects), its application and usage in the VQA task is a natural consequence. The usage of graphs instead of more typical representations (using CNNs for image feature extraction and LSTMs for word feature extraction) poses various advantages such as exploitation of the unordered nature of scene elements as well as the semantic relationships between them.

To exploit these advantages, \cite{teney2017graph} proposes a novel technique to solve this task by representing both the question and the image given by the user in the form of graphs. The training dataset used, being a synthetic one, already has object label information and their relationships for the scene graph to be a direct input, while the given question was parsed to form a graph. Thereafter, simple embeddings are generated for both nodes and edges of both the graphs and the features are then refined iteratively using a GRU. At the same time, pre-GRU embeddings are used to generate attention weights, which are used to align specific words in the question with particular elements of the scene. Once the attention weights and refined features are obtained, the weights are applied to the corresponding pairwise combinations of question and scene features. Finally with the help of non-linearities and weight vector layers, the sum of weighted features over the scene and question elements are taken step-by-step. This leads to the eventual generation of the answer vector, containing scores for the possible answers. The entire process can be seen in Fig. \ref{fig:vqa_pipeline}.

Apart from this, \cite{tang2019learning} describes a unique method of using a tree-structured representation for visual relationships. While the SGG part of the paper has already been discussed in Section \ref{methods}, it can also be used for the VQA task by changing the VCTree Construction method to incorporate pairwise task dependancy $g(x_i, x_j, q)$, alongside object correlation. The context embedding method is also altered to incorporate an additional multi-modal attention feature to calculate multi-modal joint features for question and image pair. A question guided gated decoding is proposed as well. Also, \cite{ghosh2019generating} proposes to solve the Explainable Question Answering task (explaining the reason for the answer choice made for a question) with the combined use of scene graphs and attention heatmaps to know which particular objects to look for in the scene graph.

\subsection{Image Synthesis}
The problem of image generation has been dealt in past literature with the usage of various kinds of generative models, being able to generate photo-realistic image of high quality. However, the problem of text to image synthesis \cite{reed2016generative, zhang2017stackgan} still lags behind owing to the added complexity of the generated images fitting the textual description. Just like in the case of semantic image retrieval, one of the ways to improve the model performance can be to input the description in the form of explicit relationship dependency scene graphs rather than relying on the unstructured nature of text.

\begin{figure}[t]
\captionsetup{labelfont=bf}
    \includegraphics[width=14cm]{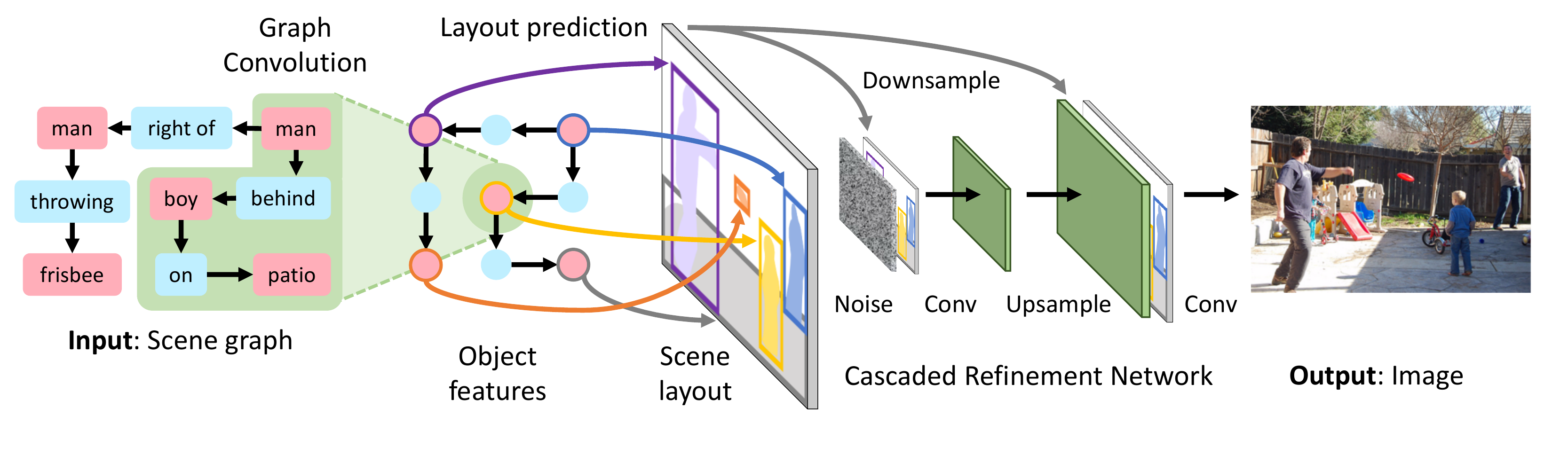}
\centering
\caption{The pipeline in \cite{johnson2018image} to use scene graphs for image generation. GCNs are used to parse the given scene graph, passing information along edges to compute embedding vectors for all objects. These vectors are used to predict the bounding boxes and segmentation masks for objects, which are combined to form a scene layout. The layout is converted into an image using \emph{Cascaded Refinement Network} (CRN) and the model is trained adversarially. During training, the model observes ground-truth object bounding boxes and (optionally) segmentation masks, but these are predicted by the model at test-time.} 
\label{fig:image_gen_pipeline}
\end{figure}

\cite{johnson2018image} proposed the first model that takes scene graph instead of text for image synthesis. The pipeline as seen in Fig. \ref{fig:image_gen_pipeline} shows the input of the model to be a scene graph which is being processed further using GCNs. Once the new object embeddings are in hand, they are further passed for scene layout creation. For this, the embeddings are passed through a \emph{mask regression network} to create a soft-binary segmentation, while an object bounding box is also generated using an \emph{object layout network}. Once the mask and box is generated, it is combined with the embedding vector to produce the object layout, and object layouts for different objects are summed to create the scene layout. For the final image creation, this scene layout is passed through \emph{Cascaded Refinement Network}, which is a series of convolutional networks with spatial resolution doubling between modules. Each module receives as input the concatenation of the scene layout (downsampled to the input resolution of the module) and the output from the previous module. Finally, the output of this module is passed through two convolutional layers to get the final image. To generate realistic images, discriminators $D_{obj}$ \& $D_{img}$ are also used for adversarial losses to generate realistic objcts and images respectively, with $D_{obj}$ also containing an auxiliary classifier to ensure that each generated object can be classified by it.

To further improve on this work, \cite{ashual2019specifying} makes one of the most evident changes, separating the layout embedding from the appearance embedding to allow much more control and freedom over object selection mechanism. Apart from that, perceptual losses are added to better capture the appearance of an object. Addition of stochasticity before mask creation  enables generation of multiple scene graphs per image.

\subsection{Image Captioning}

Image captioning is the process of generating textual descriptions from an image. The process has moved from rule and template-based approaches to CNN and RNN based frameworks. There have been striking advances in the application of  deep learning for image captioning, one of them being the use of scene graphs for image captioning. Scene graphs being a dense semantic representation of images have been used in several ways to accomplish the task of caption generation. MSDN \cite{li2017scene} proposed to solve three visual tasks in one go, namely region captioning, scene graph generation, and object detection. It offered a joint message passing scheme so that each vision task can share information and hence leveraged the rich semantic representation capability of scene graphs for image captioning.

From common understanding, it’s quite evident that modeling relationships between objects would help represent and eventually describe an image. Still, the field was unexplored until \cite{yao2018exploring} used scene graph representation of an image to encode semantic information. The idea is to develop semantic and spatial graphs and encode them as feature vectors using GCN, so that the mutual correlations or interactions between objects are the natural basis for describing an image and caption generation.

A general problem with encode-decoder based caption generation is the gap between image and sentence domains. Since scene graphs can be generated from both text and images, they can be adopted as an explicit representation to bridge the gap between the two domains \cite{gu2019unpaired,yang2019auto}. In \cite{gu2019unpaired}, the main idea was to use inductive bias as humans do, i.e., when we see the relation “person on bike,” it is natural to replace “on” with “ride” and infer “person riding bike on a road” even if the “road” is not evident. To incorporate such biases, a dictionary “$D$” based approach is adopted which is learned through the self-reconstruction of a sentence using scene graph encoder and decoder, i.e., they first generate a scene graph from the sentence \cite{anderson2016spice}, encode it using an encoder and then using the dictionary, refine the encoded features to obtain the sentence back. In the vision domain, a scene graph is constructed from an image \cite{zellers2018neural} and encoded using another encoder, after which the learned dictionary is used to refine the features which can be decoded to provide captions. 

Most works in image captioning generally deal with English captions; one major reason for that is the availability of the dataset. To overcome the problem of unavailability of paired data, \cite{yang2019auto} provides a scene graph based approach for unpaired image captioning. Everything starts with the graph construction for both image \cite{zellers2018neural} and caption \cite{anderson2016spice}. Then with the help of the same encoder, both of these graphs are converted into feature maps. The features are aligned using a GAN-based method, and finally passed through a decoder for caption generation. The use of scene graphs in \cite{gu2019unpaired,yang2019auto} proves the usability of scene graphs as a bridge between vision and text domain. 

\subsection{Evaluation Metric}
Improving on the previous n-gram matching-based approaches for automatic evaluation of Image Captioning \cite{papineni2002bleu,vedantam2015cider}, \emph{Semantic Propositional Image Caption Evaluation} (SPICE) \cite{anderson2016spice} was introduced to automatically measure caption quality based on \emph{semantic propositional content}. 

In SPICE, the candidate and reference captions are converted into scene graphs using a two stage approach. First, a Probabilistic Context-Free Grammar (PCFG) dependency parser  is used to find syntactic dependencies between words in the captions, followed by using nine simple linguistic rules to extract lemmatized objects, relations and attributes to construct a scene graph. Then, a set of logical tuples from the scene graph representing objects/attributes and relationships are extracted, which is followed by using an F1 score based approach between the candidate and reference tuple sets to calculate the SPICE metric.

SPICE shows higher correlation with human judgement than previously mentioned approaches and can also be used to evaluate more detailed subtasks like \emph{which caption generator detects people the best} by easily subdividing the tuple sets into meaningful categories. However, a potential drawback of SPICE is that it can be \emph{gamed} by a model that focuses only on object, attributes, and relationship prediction, and thus have a similar scene graph to the reference but ignore the other important aspects like grammar and syntax of the caption. Hence, we can see that the usage of scene graph \emph{does not only limit to various visual relationship tasks}, but can easily be extended to target various problems.

%% file: comparison.tex
\section{Performance Comparison}
\label{performance}

In this section, we are going to focus on quantitatively comparing the various SGG techniques described so far and also have a thorough comparative analysis of their performances. We will also define some common terms used for comparing these models. Firstly, we will be defining the various metrics commonly used in the literature, and also various newer alterations to them to make them more accurate. Secondly, we will be comparing the various models on the basis of these common metrics and notations.
\vspace{5mm}

\textbf{\emph{Term Definitions}} ~~ Metrics are a measure of quantitative assessment and need to be well understood before using them for comparison. Following the same policy, we specify the metrics used and also the parameters for comparing several SGG approaches and architectures. Rather than \emph{accuracy} or \emph{precision}, \emph{Recall} has become the de-facto choice as the model evaluation metric for the SGG task, owing to the presence of sparse annotations in the task's dataset. The notation for recall that is followed is \emph{Recall@k} abbreviated as \emph{R@k}, which has become a default choice for comparison across various methods. The \emph{R@k} metric measures the fraction of ground-truth relationship triplets (subject-predicate-object) that appear among the top-k most confident triplet predictions in an image, with the values of k being 50 and 100.

SGG is a complex task and mostly evaluated on three parameters separately as described below:
\begin{enumerate}
    \item The \emph{predicate classification} (PredCls) task is to predict the predicates of all pairwise relationships of a set of localized objects. This task examines the model’s performance on predicate classification in isolation from other factors. 
    
    \item The \emph{scene graph classification} (SGCls) or \emph{phrase classification} (PhrCls) task is to predict the predicate as well as the object categories of the subject and the object in every pairwise relationship given a set of localized objects.
    
    \item The \emph{scene graph generation} (SGGen) task is to simultaneously detect a set of objects and predict the predicate between each pair of the detected objects. An object is considered to be correctly detected if it has at least 0.5 IoU overlap with the ground-truth bounding box.
\end{enumerate}
While almost all the SGG techniques report their results on these three parameters, these evaluation metrics are far from perfect and hence some modifications have been suggested in the past literature. One of them is \emph{SGGen+} \cite{yang2018graph}, which proposes to not only consider the triplets counted by SGGen, but also the recall of singletons and pairs (if any) ,i.e., it doesn't penalize small mismatches much. For example, for a ground truth triplet $<boy-wearing-shoes>$, if $boy$ is replaced with $man$, the SGGen score will be zero due to triplet mismatching, whereas in the case of SGGen+ it will be non-zero because of the correct classification of atleast the predicate and object. Another modification suggested is that of \emph{Mean Recall@k}, abbreviated as \emph{mR@k} \cite{chen2019knowledge, tang2019learning} which tries to counter the problem of Long-tailed Dataset Distribution discussed earlier in Section \ref{methods}. If one method performs well on several most frequent relationships, it can achieve a high \emph{R@k} score, which makes it insufficient to measure performance on all relationships. This is where \emph{mean R@k} shines, as it first computes the \emph{R@k} for samples of each relationship and then averages over all relationships to obtain \emph{mR@K}, and hence giving a more comprehensive performance evaluation for all relationships.




However, \emph{we don't use SGGen+ and Mean Recall@K} in Table \ref{tab:performce_comparison} because the performance results of various models on these metrics are not available.
\vspace{5mm}

\textbf{\emph{Comparison}} ~~ The performance of the various SGG methods on Visual Genome \cite{krishna2017visual}, measured using the metrics described above, can be seen in Table \ref{tab:performce_comparison}. As can be clearly seen, the performance of all the methods with no-graph constraint is much better than ones with a graph constraint. Also, a clear decrease in the recall values as we move from PredClS to SGCls to SGGen indicates an eventual increase in the complexity of the respective metrics. While the earlier techniques such as iterative message passing \cite{xu2017scene} and MSDN \cite{li2017scene} may have laid the groundwork for further work, their lower performance can clearly be attributed to their rudimentary message passing scheme. Even with a meagre improvement in accuracy reported by Graph R-CNN \cite{yang2018graph} and Factorizable Net \cite{li2018factorizable}, they are still able to reduce their inference time, with Factorizable Net reporting upto $3 \times$ speed boost as compared to earlier techniques. The poor performance by \cite{gu2019scene, chen2019scene, dornadula2019visual} can be explained by the fact that the usage of external knowledge, semi-supervised learning and few-shot learning gives a huge bump to the accuracy only when there is a limited annotated dataset. Hence, while all these three techniques outperform the other techniques on very less amount of data, their methodology does not lead to that big of a performance boost when dealing with the whole dataset. The clear winner in this \emph{"performance race"} is evidently the usage of a modified loss function \cite{zhang2019graphical, chen2019counterfactual} and incorporating statistical dependencies \cite{zellers2018neural, chen2019knowledge, tang2019learning} in the model.

However, one thing to keep in mind here is the fact that many of these techniques used only \emph{the most frequent relationships} found in the dataset. This clearly is detrimental for the various downstream tasks that a scene graph can be used for and severely limits its applicability on real-world problems. Also, the usage of \emph{Recall@k} instead of \emph{mean Recall@k} severely handicaps the model's performance and incorporates a bias in the end result as has been described earlier. The technique proposed in \cite{tang2020unbiased}, which had the main contribution of more diverse predictions by removing the bias in predictions, report its results using only \emph{mean R@k} (which is also why we were not able to report its results in the main table) and also \emph{beats all the other models in this evaluation metric}. Furthermore, the other techniques using semi-supervised and few-shot learning methods \cite{chen2019scene, dornadula2019visual}, while performing poorly on \emph{R@k}, would see a big bump in their accuracy when using \emph{mean R@k} because of their main target being the tail classes, and have even reported to perform much better if the performance on just these infrequent classes is considered. Apart from this, the values reported in the table clearly indicate that the performance in scene graph generation is still far from human-level performance.
\begin{NoHyper} 
\begin{figure}
\centering
\captionsetup{labelfont=bf}
\begin{subfigure}{\textwidth}
\begin{tikzpicture}
\begin{axis}[
    width = 15cm,
    height = 5cm,
    ybar,
    legend style={at={(0.5,1.2)},
      anchor=north,legend columns=-1},
    symbolic x coords={MSP \cite{xu2017scene},MSP+ \cite{xu2017scene, zellers2018neural},MSDN \cite{li2017scene},FactorNet \cite{li2018factorizable},Motifs \cite{zellers2018neural},GRCNN \cite{yang2018graph},KERN \cite{chen2019knowledge},GCL \cite{zhang2019graphical},VCTree \cite{tang2019learning},ExtKnw \cite{gu2019scene},LmtdLbls \cite{chen2019scene},CCMT \cite{chen2019counterfactual},VRF \cite{dornadula2019visual}},
    xtick=data,
      x tick label style=
        {rotate=90,anchor=east}],
        
\addplot coordinates 
{
(MSP \cite{xu2017scene},3.44)
(MSP+ \cite{xu2017scene, zellers2018neural},20.7)
(MSDN \cite{li2017scene},7.73)
(FactorNet \cite{li2018factorizable},13.06)
(Motifs \cite{zellers2018neural},27.2)
(GRCNN \cite{yang2018graph},11.4)
(KERN \cite{chen2019knowledge},27.1)
(GCL \cite{zhang2019graphical},28.3)
(VCTree \cite{tang2019learning},27.9)
(ExtKnw \cite{gu2019scene},13.65)
(LmtdLbls \cite{chen2019scene},18.69)
(CCMT \cite{chen2019counterfactual},27.9)
(VRF \cite{dornadula2019visual},13.18)
};
\addplot coordinates 
{
(MSP \cite{xu2017scene},4.24)
(MSP+ \cite{xu2017scene, zellers2018neural},24.5)
(MSDN \cite{li2017scene},10.51)
(FactorNet \cite{li2018factorizable},16.47)
(Motifs \cite{zellers2018neural},30.3)
(GRCNN \cite{yang2018graph},13.7)
(KERN \cite{chen2019knowledge},29.8)
(GCL \cite{zhang2019graphical},32.7)
(VCTree \cite{tang2019learning},31.3)
(ExtKnw \cite{gu2019scene},17.57)
(LmtdLbls \cite{chen2019scene},19.28)
(CCMT \cite{chen2019counterfactual},31.2)
(VRF \cite{dornadula2019visual},13.45)
};
\legend{R@50,R@100}
\end{axis}
\end{tikzpicture}
    \caption{Comparison of methods over the SGGen metric.}
    \label{fig:my_label}
\end{subfigure}
\newline
\newline
\begin{subfigure}{\textwidth}
\centering
\begin{tikzpicture}
\begin{axis}[
    width = 15cm,
    height = 5cm,
    ybar,
    symbolic x coords={MSP \cite{xu2017scene},MSP+ \cite{xu2017scene, zellers2018neural},MSDN \cite{li2017scene},FactorNet \cite{li2018factorizable},Motifs \cite{zellers2018neural},GRCNN \cite{yang2018graph},KERN \cite{chen2019knowledge},GCL \cite{zhang2019graphical},VCTree \cite{tang2019learning},ExtKnw \cite{gu2019scene},LmtdLbls \cite{chen2019scene},CCMT \cite{chen2019counterfactual},VRF \cite{dornadula2019visual}},
    xtick=data,
      x tick label style=
        {rotate=90,anchor=east}],
        
\addplot coordinates 
{
(MSP \cite{xu2017scene},21.72)
(MSP+ \cite{xu2017scene, zellers2018neural},34.6)
(MSDN \cite{li2017scene},19.30)
(FactorNet \cite{li2018factorizable},22.84)
(Motifs \cite{zellers2018neural},35.8)
(GRCNN \cite{yang2018graph},29.6)
(KERN \cite{chen2019knowledge},36.7)
(GCL \cite{zhang2019graphical},36.8)
(VCTree \cite{tang2019learning},38.1)
(ExtKnw \cite{gu2019scene},23.51)
(LmtdLbls \cite{chen2019scene},21.34)
(CCMT \cite{chen2019counterfactual},39.0)
(VRF \cite{dornadula2019visual},23.71)
};
\addplot coordinates 
{
(MSP \cite{xu2017scene},24.38)
(MSP+ \cite{xu2017scene, zellers2018neural},35.4)
(MSDN \cite{li2017scene},21.82)
(FactorNet \cite{li2018factorizable},28.57)
(Motifs \cite{zellers2018neural},36.5)
(GRCNN \cite{yang2018graph},31.6)
(KERN \cite{chen2019knowledge},37.4)
(GCL \cite{zhang2019graphical},36.8)
(VCTree \cite{tang2019learning},38.8)
(ExtKnw \cite{gu2019scene},30.04)
(LmtdLbls \cite{chen2019scene},21.44)
(CCMT \cite{chen2019counterfactual},39.8)
(VRF \cite{dornadula2019visual},24.66)
};
\end{axis}
\end{tikzpicture}
    \caption{Comparison of methods over the SGCls metric.}
    \label{fig:my_label}
\end{subfigure}
\newline
\newline
\begin{subfigure}{\textwidth}
\centering
\begin{tikzpicture}
\begin{axis}[
    width = 15cm,
    height = 5cm,
    ybar,
    symbolic x coords={MSP \cite{xu2017scene},MSP+ \cite{xu2017scene, zellers2018neural},MSDN \cite{li2017scene},Motifs \cite{zellers2018neural},GRCNN \cite{yang2018graph},KERN \cite{chen2019knowledge},GCL \cite{zhang2019graphical},VCTree \cite{tang2019learning},LmtdLbls \cite{chen2019scene},CCMT \cite{chen2019counterfactual},VRF \cite{dornadula2019visual}},
    xtick=data,
      x tick label style=
        {rotate=90,anchor=east}],
        
\addplot coordinates 
{
(MSP \cite{xu2017scene},44.75)
(MSP+ \cite{xu2017scene, zellers2018neural},59.3)
(MSDN \cite{li2017scene},63.12)
(Motifs \cite{zellers2018neural},65.2)
(GRCNN \cite{yang2018graph},54.2)
(KERN \cite{chen2019knowledge},65.8)
(GCL \cite{zhang2019graphical},68.4)
(VCTree \cite{tang2019learning},66.4)
(LmtdLbls \cite{chen2019scene},47.04)
(CCMT \cite{chen2019counterfactual},66.4)
(VRF \cite{dornadula2019visual},56.65)
};
\addplot coordinates 
{
(MSP \cite{xu2017scene},53.08)
(MSP+ \cite{xu2017scene, zellers2018neural},61.3)
(MSDN \cite{li2017scene},66.41)
(Motifs \cite{zellers2018neural},67.1)
(GRCNN \cite{yang2018graph},59.1)
(KERN \cite{chen2019knowledge},67.6)
(GCL \cite{zhang2019graphical},68.4)
(VCTree \cite{tang2019learning},68.1)
(LmtdLbls \cite{chen2019scene},47.53)
(CCMT \cite{chen2019counterfactual},68.1)
(VRF \cite{dornadula2019visual},57.21)
};
\end{axis}
\end{tikzpicture}
    \caption{Comparison of methods over the PRDCls metric.}
    \label{fig:my_label}
\end{subfigure}
\caption{A visual representation for a performance comparison between the various SGG techniques with graph constraint on the three metrics SGGen, PhrCls and PredCls (taken from Table \ref{tab:performce_comparison}).}
\label{fig:visual_comparison}
\end{figure}
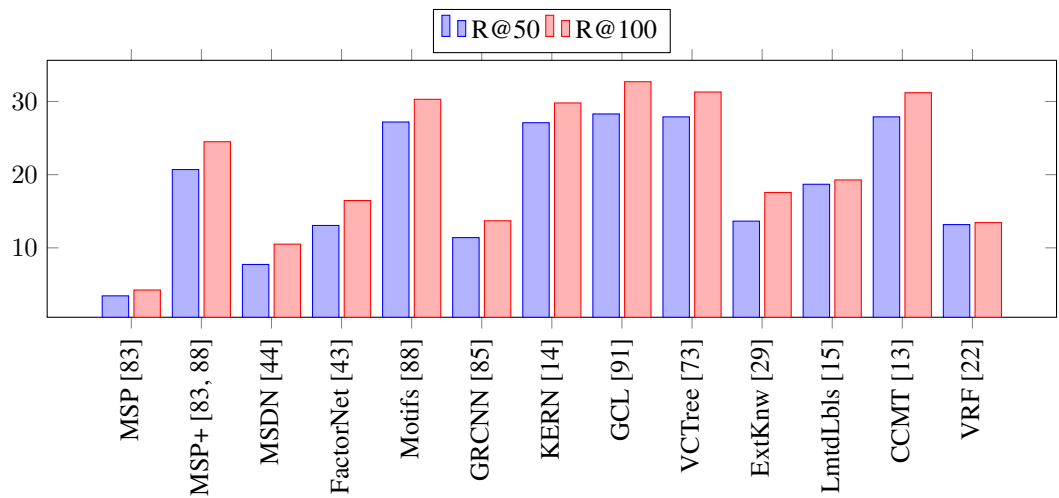
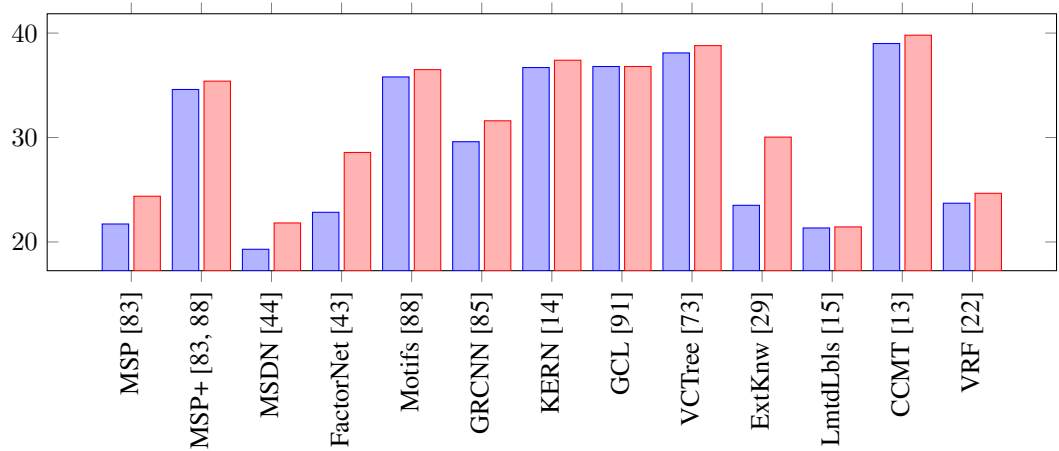
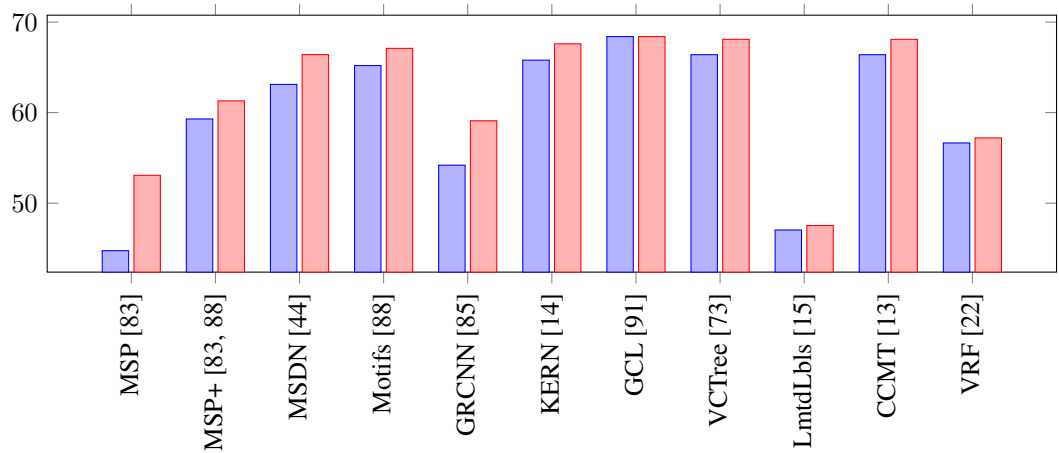

\end{NoHyper}

\begingroup
\setlength{\tabcolsep}{6pt} 
\renewcommand{\arraystretch}{1.5}
\captionsetup[table]{skip=10pt}
\begin{table}[]
\captionsetup{labelfont=bf}
\centering
\resizebox{\textwidth}{!}{%
\begin{tabular}{|l|cc|cc|cc|cc|cc|cc|}
\hline
\multicolumn{1}{|c|}{\textbf{}} & \multicolumn{6}{c|}{\textbf{Graph Constraint}} & \multicolumn{6}{c|}{\textbf{No-Graph Constraint}} \\ \cline{2-13}
\multicolumn{1}{|c|}{\textbf{Model}} & \multicolumn{2}{c|}{\textbf{SGGen}} & \multicolumn{2}{c|}{\textbf{PhrCls / SGCls}} & \multicolumn{2}{c|}{\textbf{PredCls}} & \multicolumn{2}{c|}{\textbf{SGGen}} & \multicolumn{2}{c|}{\textbf{PhrCls / SGCls}} & \multicolumn{2}{c|}{\textbf{PredCls}} \\
 & R@50 & R@100 & R@50 & R@100 & R@50 & R@100 & R@50 & R@100 & R@50 & R@100 & R@50 & R@100 \\ \hline
Message Passing \cite{xu2017scene} ( MSP ) & 3.44 & 4.24 & 21.72 & 24.38 & 44.75 & 53.08 & - & - & - & - & - & - \\
Message Passing+ \cite{xu2017scene,zellers2018neural} ( MSP+ ) & 20.7 & 24.5 & 34.6 & 35.4 & 59.3 & 61.3 & 22.0 & 27.4 & 43.4 & 47.2 & 75.2 & 83.6 \\
MSDN \cite{li2017scene} & 7.73 & 10.51 & 19.30 & 21.82 & 63.12 & 66.41 & - & - & - & - & - & - \\
Factorizable Net\footnotemark[2] \cite{li2018factorizable} ( FactorNet ) & 13.06 & 16.47 & 22.84 & 28.57 & - & - & - & - & - & - & - & - \\
Neural Motifs \cite{zellers2018neural} ( Motifs ) & 27.2 & 30.3 & 35.8 & 36.5 & 65.2 & 67.1 & 30.5 & 35.8 & 44.5 & 47.7 & 81.1 & 88.3 \\
Graph R-CNN \cite{yang2018graph} ( GRCNN ) & 11.4 & 13.7 & 29.6 & 31.6 & 54.2 & 59.1 & - & - & - & - & - & - \\
\begin{tabular}[c]{@{}l@{}}Knowledge-embedded\\ Routing Network \cite{chen2019knowledge} ( KERN ) \end{tabular} & 27.1 & 29.8 & 36.7 & 37.4 & 65.8 & 67.6 & 30.9 & 35.8 & 45.9 & 49.0 & 81.9 & 88.9 \\
Graph Contrastive Losses \cite{zhang2019graphical} ( GCL ) & 28.3 & 32.7 & 36.8 & 36.8 & 68.4 & 68.4 & 30.4 & 36.7 & 48.9 & 50.8 & 93.8 & 97.8 \\
VCTree \cite{tang2019learning} & 27.9 & 31.3 & 38.1 & 38.8 & 66.4 & 68.1 & - & - & - & - & - & - \\
\begin{tabular}[c]{@{}l@{}}External Knowledge and\\ Image Reconstruction\footnotemark[2] \cite{gu2019scene} ( ExtKnw ) \end{tabular} & 13.65 & 17.57 & 23.51 & 30.04 & - & - & - & - & - & - & - & - \\
\begin{tabular}[c]{@{}l@{}}Prediction with Limited\\ Labels \cite{chen2019scene} ( LmtdLbls ) \end{tabular} & 18.69 & 19.28 & 21.34 & 21.44 & 47.04 & 47.53 & - & - & - & - & - & - \\
\begin{tabular}[c]{@{}l@{}}Counterfactual Critic\\ Multi-agent Training \cite{chen2019counterfactual} ( CCMT ) \end{tabular} & 27.9 & 31.2 & 39.0 & 39.8 & 66.4 & 68.1 & 31.6 & 36.8 & 48.6 & 52.0 & 83.2 & 90.1 \\
\begin{tabular}[c]{@{}l@{}}Visual Relationships as\\ Functions \cite{dornadula2019visual} ( VRF ) \end{tabular} & 13.18 & 13.45 & 23.71 & 24.66 & 56.65 & 57.21 & - & - & - & - & - & - \\ \hline
\end{tabular}
}
\caption{Comparison on Visual Genome\textsuperscript{3} \cite{krishna2017visual}. Results in the table are presented in their original work setting. Graph constraints refer to a single predicate prediction per object pair whereas, in the no-graph constraint setting, multiple predicates can be detected for each object pair.}
\label{tab:performce_comparison}
\end{table}
\footnotetext[2]{Results are as per MSDN\cite{li2017scene} cleansed
dataset, whereas rest are evaluated on Visual Genome test set \cite{krishna2017visual}.}
\footnotetext[3]{There have been various different splits of the Visual Genome Dataset used in different works, making a proper comparison difficult. We have, however, tried to be as consistent as possible.}

%% file: future_directions.tex
\section{Future Directions}
\label{future_directions}

While the model performance for scene graph generation and its subsequent usage in various downstream tasks has come a long way, there is still much work that needs to be done. In this section, we will be outlining the current trend of research in this field and also some of the possible directions in which it may lead to in the future.

Starting off, as can be seen from the previous section, there has been a shift in the usage of \emph{mean Recall@k} instead of \emph{Recall@k}, which can lead to a more "reliable" evaluation of models. The long-tailed distribution in existing datasets is undoubtedly a major problem that has been targeted in more and more recent works. Some of the recent techniques used meta-learning for long-tailed recognition \cite{liu2019large} and a decoupled learning procedure \cite{kang2019decoupling}, all of which can be further modified to be used for the long-tailed distribution problem in the context of relationship detection. One of the major reasons for the limited usage of Scene Graphs for downstream tasks like VQA, Image Captioning, etc., even with such rich representational power, has been due to the handicap imposed upon them by a biased dataset. Hence, targetting the \emph{long-tail} of infrequent relationships for a more diverse prediction set has become increasingly important.

Another problem restricting the wide applicability of scene graphs is the lack of efficiency in training, which is due to the \emph{quadratic number} of candidate relations between the detected objects. With only a few of the recent techniques \cite{li2018factorizable, yang2018graph} focussing on this, we believe that much more work can be done on this front. Expanding on the application areas of scene graphs, some recent techniques \cite{armeni20193d, gay2018visual} develop a framework for utilizing the graph structure for 3D scene representation and construction, while the newly introduced Action Genome dataset \cite{ji2019action} opens up the possibility of incorporating a spatio-temporal component to the scene graphs.

With such exciting avenues still left to explore in this field, we firmly believe that Scene Graphs will be the next stepping stone for further narrowing down the gap between vision and language related tasks.






%% file: conclusion.tex
\section{Conclusion}
\label{conclusion}

This paper presented a comprehensive survey on the various techniques that have been used in the past literature for the Scene Graph Generation task. We also outlined the various problems faced in this task and how different techniques focus on solving these problems to get a more accurate model. We reviewed the various datasets common to this task as well as how scene graphs have been used so far to solve various downstream tasks. We then compared the various SGG techniques and finally discussed the possible future directions of research in this field. With such a rich representational power, we believe that scene graphs can undoubtedly lead to state-of-the-art results in various vision and language domain tasks. Seeing the increasing research work in this field for getting more accurate and diverse scene graphs, this belief of ours should become a reality sooner than later.